\documentclass[acmsmall]{acmart}

\AtBeginDocument{
  }

\setcopyright{cc}
\setcctype{by}

\copyrightyear{xxxx}
\acmYear{xxxx}
\acmDOI{xxxxx/xxxxx}

\acmJournal{THRI}
\acmVolume{xx}
\acmNumber{x}
\acmArticle{xxx}
\acmMonth{9}

\usepackage{algorithm}
\usepackage{algorithmic}
\usepackage{tabularx}
\usepackage{multirow}
\usepackage{xcolor}
\usepackage{colortbl}

\begin{document}

\title{WSM-Aware HRI: An IoT-Enhanced Framework for Early Detection and Norm-Guided Repair of Failures with LLM Guidance}

\author{Hanlin Zhang}
\email{hanlinzhang@link.cuhk.edu.cn}
\orcid{0009-0007-0349-827X}
\affiliation{
  \institution{School of Artificial Intelligence, The Chinese University of Hong Kong - Shenzhen}
  \city{Shenzhen}
  \state{Guangdong}
  \country{China}
}

\author{Yuquan Wang}
\email{haas@x2robot.com}
\orcid{0000-0002-4032-4830}
\affiliation{
  \institution{X SQUARE Robot}
  \city{Shenzhen}
  \state{Guangdong}
  \country{China}
}

\author{Tianwei Zhang}
\email{zhangtianwei@cuhk.edu.cn}
\orcid{0000-0002-1462-5402}
\authornote{Co-corresponding author.}
\affiliation{
  \institution{Shenzhen Institute of Artificial Intelligence and Robotics for Society}
  \city{Shenzhen}
  \state{Guangdong}
  \country{China}
}

\author{Zhenglong Sun}
\email{sunzhenglong@cuhk.edu.cn}
\correspondingauthor
\orcid{0000-0002-8135-1659}
\affiliation{
  \institution{School of Science and Engineering and School of Artificial Intelligence, The Chinese University of Hong Kong - Shenzhen, and Shenzhen Institute of Artificial Intelligence and Robotics for Society}
  \city{Shenzhen}
  \state{Guangdong}
  \country{China}
}

\thanks{This research was supported by the 
\grantsponsor{}{Guangdong Basic and Applied Basic Research Foundation}{} Grant No.~\grantnum{}{2026A1515011038} and
\grantsponsor{}{Shenzhen Science and Technology Program}{} under Grant No.~\grantnum{}{JSGGKQTD20221101115656029} and No.~\grantnum{}{KJZD20230923113801004}.}

\renewcommand{\shortauthors}{Zhang et al.}

\begin{abstract}
\noindent\rule{\linewidth}{0.4pt}

Human--robot interaction (HRI) failures remain a major barrier to deploying robots in real-world environments. Prior work often treats failures as isolated technical faults or focuses on post-hoc recovery behaviors. In practice, many breakdowns arise because humans and robots operate under inconsistent assumptions about the current world state. We propose WSM-Aware HRI, an IoT-enhanced modular framework that unifies diverse HRI breakdowns as \emph{World-State Mismatches (WSMs)} between a human’s instruction-implied assumptions and a robot’s grounded world model built from multimodal perception and digital augmentation. A Large Language Model (LLM) is used to make implicit assumptions explicit, map them to a small set of mismatch types, and specify the evidence needed for verification against the robot’s world state. WSM-Aware HRI shifts failure handling from execution-time recovery to proactive mismatch detection during intention formation, enabling interventions guided by safety, norm compliance, and multi-user coordination with transparent explanations. We evaluate mismatch identification in ten everyday cases spanning both visual and latent-state mismatches. The system can accurately produce the expected output results, and ablations show that reliable identification depends on appropriate grounding representations and verification-oriented refinement. These results indicate that treating interaction breakdowns as explicit world–state mismatches enables earlier detection of impending failures and offers a principled mechanism for integrating external evidence and social constraints into human–robot interaction.
\end{abstract}

\begin{CCSXML}
<ccs2012>
   <concept>
       <concept_id>10003120.10003121</concept_id>
       <concept_desc>Human-centered computing~Human computer interaction (HCI)</concept_desc>
       <concept_significance>500</concept_significance>
       </concept>
   <concept>
       <concept_id>10010520.10010553.10010554</concept_id>
       <concept_desc>Computer systems organization~Robotics</concept_desc>
       <concept_significance>500</concept_significance>
       </concept>
   <concept>
       <concept_id>10003120.10003121.10003124</concept_id>
       <concept_desc>Human-centered computing~Interaction paradigms</concept_desc>
       <concept_significance>300</concept_significance>
       </concept>
   <concept>
       <concept_id>10010147.10010178</concept_id>
       <concept_desc>Computing methodologies~Artificial intelligence</concept_desc>
       <concept_significance>300</concept_significance>
       </concept>
 </ccs2012>
\end{CCSXML}

\ccsdesc[500]{Human-centered computing~Human computer interaction (HCI)}
\ccsdesc[500]{Computer systems organization~Robotics}
\ccsdesc[300]{Human-centered computing~Interaction paradigms}
\ccsdesc[300]{Computing methodologies~Artificial intelligence}

\keywords{Human–robot interaction, world-state mismatch, proactive failure detection, large language models (LLMs), multimodal perception, IoT augmentation, context-aware robotics, norm-guided repair, explainable robotics, collaborative recovery}

\received{}
\received[revised]{xx xx}
\received[accepted]{xx xx}

\maketitle

\section{Introduction}

Errors and failures in Human--Robot Interactions (HRI) remain a central barrier to moving robots from controlled laboratory demonstrations to robust real-world deployment in homes, clinics, schools, and public service environments \citep{Honig18}. Unlike industrial environments with well-defined procedures, daily human-Robot interaction is open and changing over time \citep{Tolmeijer20}. The instructions from users are often not clear enough and rely on implicit common-sense assumptions \citep{Deits13,Tellex11}. During the interaction process, users with different positions may propose conflicting goals and constraints \citep{Tolmeijer20}. In such settings, even successful task execution can still constitute an interaction failure if it violates safety expectations, institutional rules, or social norms \citep{Tian21,Mirnig17}.

A key reason is that many breakdowns are not reducible to isolated technical faults, nor are they fully addressed by post-hoc conversational recovery \citep{Tolmeijer20,Reig21}. Instead, interaction failures often emerge from a deeper coordination problem that humans and robots act on different beliefs about the world state and its constraints. 
This view is related to the notion of common ground in dialogue and situated interaction, where successful coordination depends on participants maintaining sufficiently aligned assumptions about the task, context, and referents. In embodied HRI, however, such assumptions are not merely conversational commitments. They may correspond to physical affordances and other users' goals. We therefore treat WSM as an action-oriented operationalization of common-ground breakdowns: instruction-implied assumptions are represented as verifiable world-state constraints and checked against the robot's grounded state before execution.
Building upon common-ground approaches that study how interaction partners establish and maintain shared assumptions, WSM introduces an embodied verification layer that operationalizes whether these assumptions remain consistent with the robot’s grounded world state, including perceptual, digital, and normative conditions. Thus, WSM provides a mechanism for translating implicit alignment requirements into explicit constraints, evidence checks, and intervention decisions before action execution.

A user may assume an object is safe to manipulate while the robot perceives fragility \citep{BobuHri21}. A user may also request an action that appears reasonable but conflicts with the robot’s hazard assessment \citep{Tolmeijer20}, such as using water to put out a fire that involves an electrical source. In other cases, an instruction may overlook privacy expectations, role-based permissions, or culturally grounded norms \citep{Denning09,Tian21}. These failures are often latent at the moment an instruction is issued, well before any physical execution error becomes observable, which suggests that robust HRI requires models that can proactively represent and adjust misaligned assumptions \citep{Deits13,BobuHri21}. 
Importantly, these mismatches may be latent rather than already perceived by the human. A user may issue an instruction under an incorrect assumption, while the resulting failure condition only becomes visible if the robot acts on that assumption. In this sense, proactive failure handling means verifying instruction-implied conditions during intention formation, before the robot commits to an irreversible or unsafe action. Many such checks can be performed through onboard sensing, but others require digitally augmented evidence, such as smart-device states, environmental sensors, access-control records, or institutional policies, when the relevant state is outside the robot's direct perceptual field.

Recent work has made substantial progress in understanding and detecting errors and failures using multimodal messages. Error-aware HRI systems increasingly leverage facial expressions and speech to detect robot errors and user repair attempts \cite{Stiber23}. As one of benchmark-driven efforts, ERR@HRI Challenge provides annotated multimodal features for facial, speech, and pose cues to identify robot mistakes and interaction ruptures, enabling systematic comparison of detection models on conversational and social breakdowns \cite{Spitale24}. Multimodal fusion is also effective for online execution monitoring in physical interaction. FINO-Net \cite{Inceoglu21} fuses RGB, depth and audio to detect and identify manipulation failures in unstructured environments. Their subsequent work extends this study by using a shared exteroceptive setup to detect and classify manipulation and post-manipulation failures \cite{Inceoglu24}. These studies demonstrate that failures can be recognized from multimodal messages across both social and physical interaction.

At the same time, advances in instruction-following robotics, especially for the Large Language Model employed on robots, heighten both the promise and the risk of deploying robots in high-variance real-world contexts. Work on grounding language in robotic affordances emphasizes that robots must constrain high-level instructions by what is physically feasible and contextually appropriate, rather than executing text literally \cite{Ichter23a}. 
Similarly, RT-2 transfers web-scale vision-language knowledge into robotic control, expanding semantic generalization and reasoning capabilities across robotic tasks \cite{Zitkovich23a}. However, such vision-language-action systems primarily address how to map instructions and observations to executable robot actions. They leave open a complementary HRI question: before committing to an action, how should a robot verify whether the user’s request presupposes world-state conditions, safety constraints, institutional rules, or social norms that are false, unverifiable, or inadmissible?
While recent vision-language-action (VLA) and multimodal language models aim to unify perception, reasoning, and action generation, they primarily focus on producing executable behaviors from instructions and observations. In embodied HRI, however, successful action generation does not necessarily guarantee appropriate interaction outcomes, because user instructions often contain implicit assumptions about physical states, temporal conditions, social constraints, and external context. Explicitly integrating these heterogeneous factors into an end-to-end policy remains challenging, particularly when the relevant information is distributed across robot perception, digital environments, and human-centered constraints. Therefore, an intermediate verification process that evaluates whether instruction-implied assumptions remain consistent with the current world state can provide a complementary capability to existing action-generation approaches. WSM-Aware HRI is designed from this perspective: rather than replacing instruction-following models, it introduces an interaction-level verification mechanism that supports robots in deciding whether to execute, clarify, or intervene before action commitment.

Despite rapid progress in multimodal detection and in interaction-level recovery, the literature still lacks a unified framework that (i) explains diverse HRI failures through a common mechanism, (ii) supports proactive detection at the stage of intention formation, and (iii) enables principled repair under safety, ethics, and multi-user coordination constraints. Existing multimodal analyses show that humans respond to conversational failures with systematic behavioral signatures that can be leveraged for automated detection, but they also highlight the absence of a general account of why breakdowns occur and how to prevent them early across robot embodiments and tasks \cite{Kontogiorgos20}. On the repair side, empirical evidence indicates that how a robot handles errors can shape users’ trust trajectories and that individual attitudes modulate the effectiveness of trust repair strategies \cite{Esterwood22}. Although real-world failures are often triggered by missing context outside the robot’s onboard sensors such as weather, air quality and so on, integrating external information from the Internet of Things (IoT) into a robot’s world model and using it to prevent and repair interaction failures remains underexplored. Concretely, these interaction-facing failures can be understood as world-state mismatches that fall into a small and reusable taxonomy: instruction–logic mismatches, perceptual/affordance mismatches, temporal/sequence mismatches, social/ethical norm mismatches, and multi-user conflicts.

To address these limitations, we propose WSM-Aware HRI, an IoT-enhanced and modular framework that unifies HRI failures as World-State Mismatches (WSM). This refers to the world assumptions inferred by humans from the history of language, behavior and interaction, which are systematically inconsistent with the world state models derived by robots from multimodal perception and Internet of Things. The core idea is that robots should not merely execute instructions reactively. Instead, they should proactively maintain an extended world-state representation and compare it against inferred human assumptions to detect mismatches early and resolve them via some principles to repair. 

\begin{figure}[t]
  \centering
  \includegraphics[width=\linewidth]{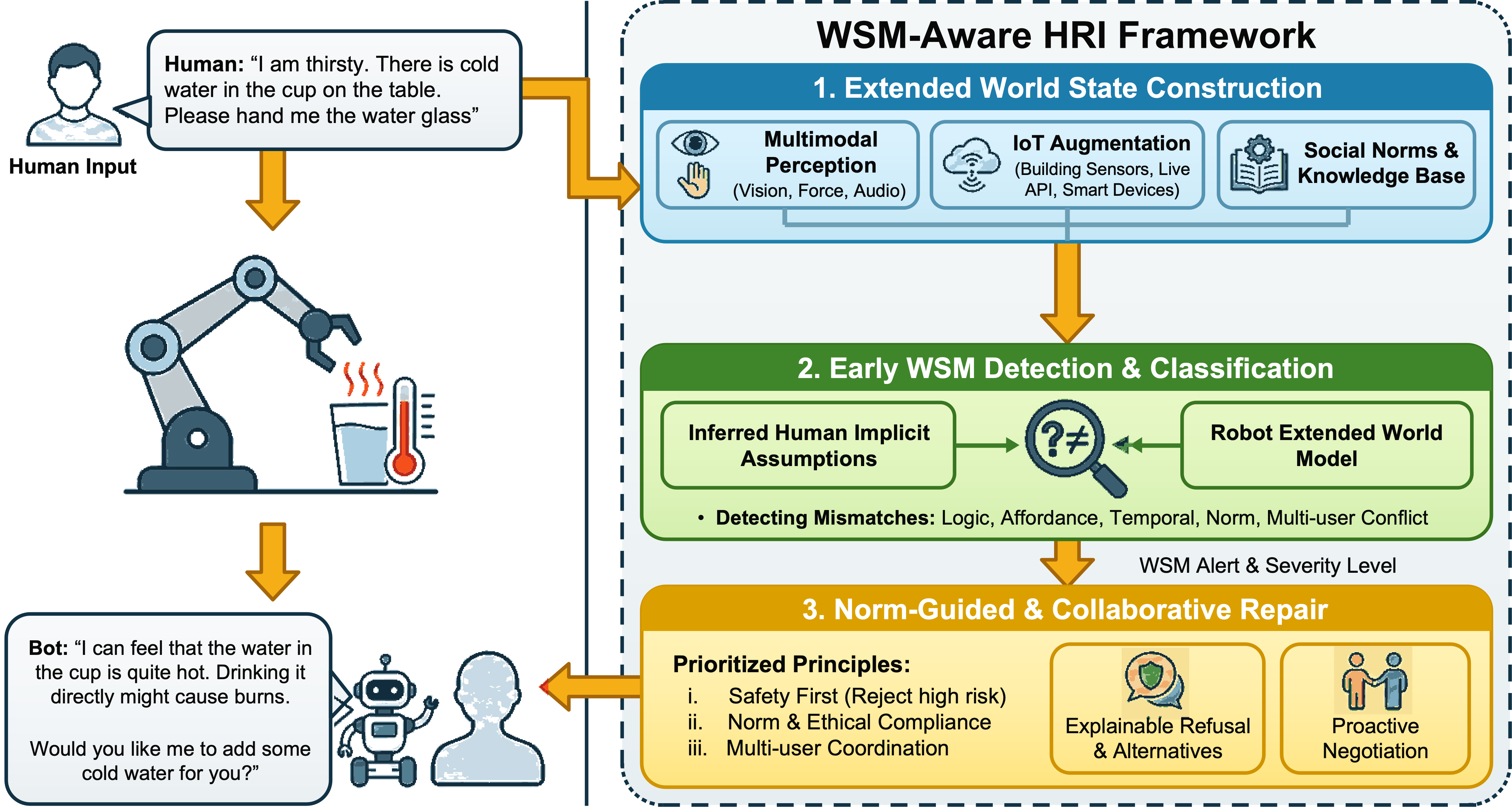}
  \caption{WSM-Aware HRI Framework.}
  \Description{}
  \label{fig:wsm-framework}
\end{figure}

Figure~\ref{fig:wsm-framework} shows the WSM-Aware HRI Framework.
We focus on instruction-following HRI episodes in which user requests presuppose conditions about the physical, temporal, normative, or multi-user state of the environment. In this setting, WSM does not replace low-level robot fault diagnosis; rather, it provides an interaction-level account of failures that arise when instruction-implied assumptions are unsupported, contradicted, or inadmissible with respect to the robot’s grounded world model. This framing allows perception, planning, dialogue, and execution issues to be analyzed when they affect the verification of assumptions required for safe and appropriate action.

The main contributions of this article are threefold. First, we propose World-State Mismatches (WSM) that define the types of failures and explain diverse HRI failures as misalignment between human assumptions and robot world models.
Second, we design an IoT-augmented world-state representation that combines multimodal sensing with external digital context to capture deployment-critical constraints. Finally, we present a norm-guided repair framework that resolves mismatches with safety-first, ethics-aware, and multi-user coordination strategies.

\section{Related Work}
\subsection{Multimodal Breakdown Detection in HRI}
More and more HRI research treats interaction breakdowns as multimodally observable coordination problems that manifest not only as task failure, but also as changes in users’ nonverbal and paralinguistic behavior. These changes include facial activity, vocal prosody, and timing irregularities that signal confusion, misalignment, or attempts to repair the interaction. A representative work analyzes conversational robot failures across datasets and tasks to identify cross-situational behavioral signatures that are predictive of breakdowns, showing that facial and acoustic features can support automatic detection beyond system-side logs \citep{Kontogiorgos21}. Building on this empirical foundation, recent modeling efforts increasingly frame breakdown detection as a time-series prediction problem over heterogeneous streams, integrating facial expressions, acoustic cues, and linguistic representations to detect conversational errors and user-initiated corrections within a unified temporal pipeline \citep{Wachowiak24,Pramanick24,Janssens24,Spitale24}.

Beyond conversational rupture labels, several multimodal datasets and analyses focus on breakdown-adjacent user states that are practically important for detection, particularly confusion and affective responses \citep{LiRoss23}. A recent dataset contribution synchronizes face and upper-body video with user speech to support modeling and detecting user confusion in situated, task-oriented HRI, explicitly distinguishing confusion-related social behaviors across scenarios \citep{LiCourtneyRoss25}. Related empirical work further analyzes how observable multimodal features correlate with graded confusion levels in task-oriented HRI and motivates social aware interaction strategies grounded in measurable behavioral cues \citep{LiRoss23}.

\subsection{Failure Monitoring and Classification in Physical HRI}
%% 2) Failure Monitoring and Classification in Physical Human–Robot Interaction
Robotics research has long studied failures in physical interaction as deviations during execution, with the goal of detecting failures early, localizing when they occur, and classifying their types. These methods rely on rich sensing, especially multimodal streams from cameras, depth sensors, microphones, force sensors, and joint signals. 
Recent work on humanoid robotic perception further highlights the role of multi-sensor perception in constructing reliable representations of the robot and its surrounding environment~\cite{bin2025survey}.
In manipulation, multimodal fusion is widely used for online failure detection in unstructured settings. FINO-Net supports both manipulation-phase and post-manipulation-phase failure detection and classification across tabletop tasks \cite{Inceoglu21,Inceoglu24}. Related work also studies failure handling under degraded perception, combining visual tracking with additional cues to improve robustness in pick-and-place tasks under partial occlusion \cite{Zhu21}.

Another direction focuses on predicting failures early enough to enable intervention. In navigation, proactive anomaly detection estimates the probability of future failures by combining planned motion with multi-sensor observations \cite{Ji22}. Other work treats failure handling as diagnosis and model correction. After a failure, robots test possible causes against parameterized execution models and update these models to reduce recurrence \cite{Mitrevski21}. PokeRRT \cite{Pasricha22} introduces poking as a non-prehensile primitive to recover from failed grasps and reachability limits.

While these approaches effectively monitor execution-level deviations, they typically define failure relative to physical outcomes or trajectories and do not explicitly account for human assumptions, task semantics, or interaction-level constraints that often trigger failures in human–robot interaction.

\subsection{Instruction-Following Robots: Action Grounding and Pre-Commitment Verification}

Recent progress in instruction-following robotics is driven by Large Language Models (LLM) and transformer-based control architectures that map natural language instructions to long-horizon action sequences. RT-1\cite{Brohan23} demonstrates a high-capacity robotics transformer trained on diverse real-robot data, enabling generalization across many everyday manipulation tasks. PaLM-E\cite{Driess23} further advances grounding by integrating continuous robot state and perceptual inputs into a large multimodal language model, supporting embodied reasoning across multiple tasks and embodiments.

Alongside end-to-end policies, a complementary family of approaches uses language models to generate intermediate programmatic structures that improve executability and enable closed-loop correction. Code as Policies\cite{Liang23} shows that code-generating language models can produce robot policy programs that compose perception modules and low-level skills into reactive behaviors. Inner Monologue \cite{Huang22} further demonstrates how language models can incorporate environment feedback during planning to improve long-horizon task completion.

These advances also sharpen a deployment risk that natural language instructions often omit critical preconditions, constraints, or social considerations, and such that literal compliance can lead to unsafe or norm-violating behavior. From an HRI perspective, underspecification frequently reflects incomplete or incorrect assumptions about the world state rather than mere ambiguity in wording. Prior work analyzes how clarification itself can introduce moral risks and proposes approaches for morally sensitive clarification requests in interactive settings \cite{Jackson22}. These findings motivate mechanisms that explicitly represent missing assumptions and reason about conflicts between user intent, sensed reality, and normative constraints.

This line of work shows how language models can help robots translate natural language into executable behavior, for example through skill selection, program synthesis, policy grounding, or closed-loop planning \cite{Liang23,Huang22,Brohan23,Zitkovich23a,Ichter23a}. WSM-Aware HRI builds on these advances but addresses a different point in the interaction pipeline. Rather than using the LLM primarily to generate or select the next action, we use it before action commitment to make instruction-implied assumptions explicit, propose candidate mismatch types, and identify the evidence needed to test those assumptions. The final mismatch judgment is therefore grounded not in the LLM output alone, but in whether the proposed assumptions are supported by the robot's perception, execution state, digitally augmented evidence, and applicable norms. In this sense, WSM-Aware HRI is complementary to systems such as SayCan, RT-1, and RT-2: while these systems primarily improve action grounding and policy generalization, our framework focuses on pre-commitment mismatch checking and norm-guided intervention before execution begins.

\subsection{Situated Assumptions in Human--Robot Interaction}

Prior work on common ground in dialogue and situated interaction shows that successful coordination depends on participants maintaining sufficiently aligned assumptions about references, task goals, and interaction context. In HRI, however, the assumptions carried by an instruction are often tied to embodied action. A request may presuppose that an object is safe to grasp, that a prerequisite action has already been completed, that the user has the authority to request an action, or that no other stakeholder has a conflicting goal. These assumptions are related to common ground, but they cannot always be resolved through linguistic alignment alone, because they may depend on physical affordances, temporal state, social permissions, or external contextual information.

WSM-Aware HRI treats these instruction-implied assumptions as conditions that should be checked before action execution. The relevant question is therefore not only whether the human and robot appear to share a conversational understanding, but whether the conditions presupposed by the instruction are supported by the robot's grounded world state. This world state may include onboard perception, execution status, digitally augmented evidence, and applicable social or institutional constraints. When such conditions are unsupported, the resulting mismatch can guide clarification, refusal, safe alternatives, or multi-user negotiation. In this way, WSM provides an action-oriented mechanism for connecting common-ground breakdowns to proactive failure detection and repair in embodied HRI.

Existing common-ground approaches primarily investigate how interaction partners establish, maintain, and repair aligned assumptions during collaboration. WSM-Aware HRI extends this line of research by addressing a complementary operational question in embodied interaction: whether the assumptions underlying an intended action remain supported by the robot’s current world state. Rather than introducing another mechanism for establishing shared beliefs, WSM operationalizes common-ground breakdowns by converting implicit assumptions into explicit constraints that can be verified through perception, IoT-derived evidence, execution state, and normative information. This enables an embodied verification and intervention layer between assumption formation and action commitment.

\subsection{Repair and Trust in HRI}
HRI research studies repair as an interactional response to breakdowns that can preserve cooperation and trust over time. Many systems employ social-account strategies such as apology, explanation, denial, and promise, and these choices shape how users attribute responsibility and predict future reliability. Experimental work shows that attributional framing matters that internal-attribution apologies can improve trust outcomes relative to denial, while denial may perform worse than providing no repair in some conditions \cite{ZhangIJHCS23}. Complementary results further demonstrate interactions between failure types and repair strategies, with logic-related failures often producing stronger trust damage and internal-attribution apology emerging as a robust baseline across conditions \cite{ZhangIJSR23}.

Repair effectiveness also depends on repetition and interaction history. As trust violations accumulate, common verbal repairs such as apologies, explanations, and promises lose their ability to restore perceived ability and integrity \cite{Esterwood23}. Other studies find that promises can become a liability. When a robot fails to keep a promise, trust drops more sharply than after simpler repair strategies, highlighting the risks of overcommitment in recovery dialogue \cite{Nesset23}. These findings suggest that effective repair cannot rely solely on surface-level strategies, but requires principled decision-making that accounts for safety, norms, and interaction history.

\subsection{Summary}
Across HRI and robotics research, prior work has made substantial progress in detecting and responding to interaction failures, yet these advances remain fragmented across problem formulations and system layers. Multimodal HRI studies demonstrate that interaction breakdowns can be reliably detected from users’ facial, vocal, and temporal behaviors, capturing confusion, misalignment, and repair attempts as they emerge in interaction \cite{Kontogiorgos21,Ruddy25}. Robotics research develops execution-level monitoring techniques that identify and classify physical failures from multimodal sensor data and can even predict failures before completion, particularly in manipulation and navigation tasks \cite{Inceoglu21,Inceoglu24,Ji22}. Meanwhile, instruction-following and language-grounded robotics significantly expand robots’ ability to interpret natural language, but also expose risks associated with underspecified instructions and missing contextual constraints, which motivates clarification and refusal mechanisms that reason about intent and moral considerations \cite{Brohan23,Driess23,Liang23,Jackson22}. HRI research on repair and trust shows that recovery strategies such as apology, explanation, and promise interact with failure type over time, and that surface-level strategies alone are insufficient for sustaining trust in repeated or high-stakes interactions \cite{ZhangIJHCS23,Esterwood23,Nesset23}. These works reveal a common limitation that failures are typically addressed at the level of observable symptoms, execution deviations, or post-hoc repair strategies. They still lack a unified representation of the mismatches between human assumptions and robot world-state perceptions that give rise to breakdowns across interaction, physical execution, and normative contexts.

\section{Methodology}
\subsection{Problem Setup}
We study a human-robot interaction episode in which a user issues an instruction $q$ and the robot executes a sequence of actions. We model many interaction failures as \emph{World-State Mismatches (WSM)}, which includes inconsistencies between (i) the world state implicitly assumed by the human and (ii) the world state maintained by the robot using multimodal sensing and digital augmentation.

This problem setup delimits the role of WSM-Aware HRI. We consider failures at the interaction level, where the relevant question is whether the robot should proceed, clarify, refuse, gather more evidence, or propose an alternative before committing to action. Low-level hardware faults, actuator degradation, controller instability, or internal software exceptions are not treated as WSMs by themselves. However, perception errors, planning inconsistencies, navigation failures, manipulation failures, and dialogue misunderstandings can become WSM-relevant when they affect the robot’s ability to verify an instruction-implied condition or when they create a discrepancy between what the user assumes and what the robot can safely or normatively execute.

Let the robot execute a high-level plan $\Pi = (a_1, a_2, \ldots, a_T)$, where each step can be either a task action or an information-gathering probe. The robot maintains a time-indexed, queryable world state and continuously monitors mismatch risk before committing to potentially unsafe or norm-violating actions, enabling early warning and proactive intervention.
In our framework, a failure condition does not require that the human has already noticed a problem. A World-State Mismatch is present whenever the instruction presupposes a condition that the human implicitly assumes to hold, but that the robot cannot verify in its grounded world state. Correspondingly, we use proactive in a pre-commitment sense that after receiving an instruction, the robot immediately externalizes its implied assumptions, evaluates admissibility, and performs targeted verification before executing the first irreversible action. The framework is thus not limited to reacting after an observed failure. It intervenes when a failure condition is already objectively present, even if it has not yet become interactionally visible.

\subsection{World-State Representations}

We represent the human's implicit assumptions as a structured latent state:
\begin{equation}
W_h \triangleq \langle I_h, C_h, N_h \rangle
\end{equation}
where $I_h$ denotes inferred intent and goal constraints, $C_h$ denotes assumed physical context, and $N_h$ denotes assumed normative context.

We infer $W_h$ from the instruction $q$ and the interaction signals observed up to time $t$:
\begin{equation}
W_h^t \leftarrow \textsc{InferHumanState}(q, u_{1:t})
\end{equation}
where $u_{1:t}$ aggregates multimodal user cues such as prosody, hesitation, gaze shifts, and corrective behaviors. In practice, $\textsc{InferHumanState}$ can be instantiated by a multimodal model and a Large Language Model that produces a structured hypothesis of human assumptions for downstream comparison and explanation.

The robot maintains an augmented world state:
\begin{equation}
W_r^t \triangleq \langle S_r^t, A_r^t, N_r^t \rangle
\end{equation}
where $S_r^t$ encodes multimodal physical sensing, $A_r^t$ encodes digital augmentation signals from IoT sources and external services, and $N_r^t$ encodes normative and role constraints available to the robot. $W_r^t$ is designed to be queryable and updateable to support auditing and transparent decision making.
In this representation, we do not assume a fixed IoT stack or a particular class of external service. Instead, heterogeneous external sources are treated as providers of verifiable updates to the augmentation component $A_r^t$. Once incorporated into the robot’s augmented world state, these observations are handled through the same admissibility checks, constraint-evaluation procedures, and mismatch-typing mechanisms as onboard evidence.

IoT augmentation is not required for every mismatch case. It becomes relevant when an instruction-implied condition cannot be reliably established from local perception or onboard execution state alone, for example when the decisive state is external, latent, future-oriented, or digitally mediated. In such cases, IoT devices and external services provide an additional evidence channel rather than a separate reasoning layer.

\subsection{WSM Discrepancy and Admissibility}

We define an admissibility predicate that indicates whether executing instruction $q$ is allowed under the current states:
\begin{equation}
\mathcal{A}(q, W_h^t, W_r^t) \in \{0,1\}
\end{equation}
$\mathcal{A}(\cdot)=1$ means the instruction is executable under safety constraints and normative constraints. $\mathcal{A}(\cdot)=0$ means the robot should intervene through clarification, refusal, or alternative suggestions. In our framework, $\mathcal{A}$ serves as a feasibility-and-norm check that can incorporate both physical safety and policy/ethics constraints.

We define a discrepancy score between human assumptions and the robot's augmented world model:
\begin{equation}
    \Delta(W_h^t, W_r^t) =
\lambda_p \Delta_p(C_h^t, S_r^t) +
\lambda_d \Delta_d(C_h^t, A_r^t) +
\lambda_n \Delta_n(N_h^t, N_r^t) +
\lambda_i \Delta_i(I_h^t, \hat{I}_r^t)
\end{equation}
where $\Delta_p$ measures physical mismatch, $\Delta_d$ measures mismatch revealed through digital augmentation, $\Delta_n$ measures normative mismatch, and $\Delta_i$ measures intent mismatch. The $\lambda$ terms weight the contributions. 

We trigger a WSM alert when mismatch risk exceeds a threshold or admissibility fails:
\begin{equation}
\textsc{TriggerWSM}(t) \Leftrightarrow
\Big(\Delta(W_h^t, W_r^t) > \tau\Big) \ \lor\ \Big(\mathcal{A}(q, W_h^t, W_r^t)=0\Big)
\end{equation}

The threshold $\tau$ can be adaptive when it should be calibrated by task risk, user trust history, or environment uncertainty, supporting proactive detection before failures fully manifest.

When a trigger occurs, we assign a mismatch type
\begin{equation}
z^t \leftarrow \textsc{ClassifyWSM}(W_h^t, W_r^t), \quad z^t \in \mathcal{Z}
\end{equation}
where $\mathcal{Z}$ includes instruction--logic mismatch, perceptual/affordance mismatch, temporal/sequence mismatch, social/ethical norm mismatch, and multi-user conflict mismatch. 

\vspace{-10pt}
\subsection{Taxonomy Construction Procedure}

\begin{table*}[t]
\centering
\caption{World-State Mismatch (WSM) Types, Definitions and Representative Examples}
\label{tab:wsm_examples}
\renewcommand{\arraystretch}{1.15}
\setlength{\tabcolsep}{6pt}

\begin{tabularx}{\textwidth}{p{3.2cm}X}
\toprule
\textbf{WSM Type} & \textbf{Definition and Representative Example} \\
\midrule

Instruction Logic Mismatch &
\textbf{Definition:} The instruction is incompatible with task-level logic or hazard/causal constraints under the current context.\newline
\textbf{Example:} Ask the robot to extinguish a fire with water while the robot detects an active electrical source.
\\
\midrule

Perceptual / Affordance Mismatch &
\textbf{Definition:} The user’s instruction assumes physical feasibility or safe affordances that are contradicted by grounded sensing or verification.\newline
\textbf{Example:} Request the robot to pick up a cup with cold water while the robot detects hot liquid inside.
\\
\midrule

Temporal / Sequence Mismatch &
\textbf{Definition:} The instruction conflicts with temporal ordering or state-machine prerequisites.\newline
\textbf{Example:} Ask the robot to place an object before it has been grasped.
\\
\midrule

Social / Ethical (Norm) Mismatch &
\textbf{Definition:} The instruction violates applicable social norms, policies, or role-based permissions, regardless of physical feasibility.\newline
\textbf{Example:} Request the robot to enter a restricted or private area.
\\
\midrule

Multi-user Conflict &
\textbf{Definition:} The instruction is incompatible with constraints or objectives posed by multiple stakeholders, requiring coordination or negotiation.\newline
\textbf{Example:} The teacher requests the robot to maintain classroom order while students request entertainment.
\\
\bottomrule
\end{tabularx}
\vspace{-10pt}
\end{table*}

Table~\ref{tab:wsm_examples} presents the five World-State Mismatch (WSM) types, their definitions, and representative examples. We organize these types around the instruction-implied constraints that a robot may need to verify before or during execution. These constraints include task-logic assumptions, physical-affordance assumptions, temporal prerequisites, social or institutional norms, and compatibility among stakeholder goals. These WSM types draw on prior work on HRI failures, robot error taxonomies, social errors, trust-relevant failures, execution monitoring, and interaction breakdowns, but reinterpret these failure mechanisms as constraints that can be verified against the robot's grounded world state.

\begin{table*}[t]
\begingroup
\centering
\caption{Mapping prior failure mechanisms to WSM categories, verification evidence, and repair responses.}
\label{tab:taxonomy_mapping}
\scriptsize
\renewcommand{\arraystretch}{1.18}
\setlength{\tabcolsep}{4pt}
\begin{tabularx}{\textwidth}{
p{0.19\textwidth}
>{\raggedright\arraybackslash}X
>{\raggedright\arraybackslash}X
>{\raggedright\arraybackslash}X
p{0.14\textwidth}}
\toprule
Prior failure mechanism &
Instruction-implied constraint &
Evidence needed for verification &
Repair implication &
Resulting WSM category \\
\midrule

Perception/action mismatch and physical execution failure~\cite{Honig18,Cameron}
& The instruction assumes that the target object or action is physically feasible, compatible, and safe under the current context.
& Grounded perception, object state, affordance checks, force/temperature/weight sensing, or other physical verification.
& Verify the relevant property, adapt the plan, avoid unsafe manipulation, or suggest a safer alternative.
& Perceptual/Affordance Mismatch \\[0.45em]

Mismatched expectations and invalid causal/task assumptions~\cite{Honig18,Salem8520654}
& The instruction presupposes a causal, task-level, or goal-logic condition that may not hold in the current world state.
& Task model, hazard relation, commonsense constraint, causal dependency, or context-specific safety rule.
& Clarify the intended goal, reject invalid causal assumptions, or propose a goal-preserving alternative.
& Instruction Logic Mismatch \\[0.45em]

Expectation-related timing, prerequisite, or ordering failure~\cite{Tolmeijer20,Cameron}
& The instruction assumes that a required prior state has already been achieved or that the requested action is temporally admissible.
& Execution history, task-state record, state-machine status, prerequisite checks, or temporal ordering constraints.
& Insert missing prerequisites, reorder the plan, delay execution, or request confirmation before proceeding.
& Temporal/Sequence Mismatch \\[0.45em]

Social error, privacy violation, permission failure, or role/policy violation~\cite{Tian21,nogueira2024taxonomy}
& The instruction assumes that the requested action is socially, ethically, or institutionally permissible.
& Role information, permission status, privacy policy, institutional rule, safety norm, or social constraint.
& Request consent or authorization, refuse the action with an explanation, or provide a norm-compliant alternative.
& Social/Ethical Norm Mismatch \\[0.45em]

Intertwined social causes and incompatible stakeholder goals~\cite{civit2025multiuser,10.1145/3415247}
& The instruction assumes that the goals or constraints of multiple users are mutually compatible.
& Active user requests, stakeholder roles, priority rules, conflict status, or multi-user interaction history.
& Initiate negotiation, arbitrate according to priority rules, escalate to a human authority, or propose a compromise.
& Multi-user Conflict \\
\bottomrule
\end{tabularx}
\endgroup
\end{table*}

To make the taxonomy operational, we first identified recurring failure mechanisms in prior work that are relevant to instruction-following HRI, such as physical infeasibility, invalid task assumptions, unmet prerequisites, privacy or permission violations, and conflicts among stakeholder goals. We then reformulated these mechanisms as constraints that can be checked against the robot's grounded world state. Finally, we grouped the constraints according to the kind of evidence needed for verification and the type of repair response implied by a violation. As a result, each WSM category corresponds not only to a descriptive failure type, but also to a checking procedure and a repair pathway in the WSM-aware pipeline.

The mapping in Table~\ref{tab:taxonomy_mapping} reflects several complementary strands of prior work. Honig and Oron-Gilad's distinction between technical failures rooted in perception/action mismatches and interaction failures arising from mismatched expectations~\cite{Honig18} motivates our separation between Perceptual/Affordance Mismatch and Instruction Logic Mismatch. Tian and Oviatt's taxonomy of social errors that undermine socio-affective competence~\cite{Tian21} motivates the treatment of privacy, permission, role, and norm-compliance violations as Social/Ethical (Norm) Mismatches. Tolmeijer et al.'s categorization of trust-relevant failures and mitigation strategies~\cite{Tolmeijer20} informs Temporal/Sequence Mismatches, where the user's instruction conflicts with prerequisite states, ordering constraints, or expected task progression. Cameron et al.'s taxonomy of domestic robot failure outcomes~\cite{Cameron} supports the inclusion of everyday household failures involving object properties, affordances, and execution prerequisites, which are reflected in the Perceptual/Affordance and Temporal/Sequence categories. Nogueira's multidimensional taxonomy of interaction failures for transparent robots, which emphasizes user-perceived breakdowns~\cite{nogueira2024taxonomy}, further supports the explicit treatment of social and normative failures. Work on HRI accident errors and multi-user interaction highlights intertwined physical and social causes~\cite{civit2025multiuser,10.1145/3415247}, supporting our distinction between Multi-user Conflict and individual norm violation, since conflicting stakeholder goals require coordination or arbitration rather than permission checking alone.

\subsection{Mismatch Types Definitions}
\label{sec:mismatch_taxonomy_typing}

Given an instruction $q$ and interaction context up to time $t$, we represent implicit assumptions required for successful and appropriate execution as a set of constraints partitioned into five families:
\begin{equation}
\mathcal{C}(q)\triangleq 
\mathcal{C}_{R}(q)\ \cup\ 
\mathcal{C}_{M}(q)\ \cup\ 
\mathcal{C}_{T}(q)\ \cup\ 
\mathcal{C}_{P}(q)\ \cup\ 
\mathcal{C}_{L}(q)
\end{equation}
where $\mathcal{C}_{R}$ denotes normative/policy constraints, $\mathcal{C}_{M}$ denotes multi-user constraints capturing compatibility of objectives across stakeholders, $\mathcal{C}_{T}$ denotes temporal constraints, $\mathcal{C}_{P}$ denotes physical preconditions and affordances, and $\mathcal{C}_{L}$ denotes instruction-level logic constraints implied by task semantics.

We evaluate any constraint $\phi\in\mathcal{C}(q)$ under both the estimated human-assumption state $W_h^t$ and the robot's augmented world state $W_r^t$:
\begin{equation}
\text{sat}_h(\phi)\triangleq \mathbf{1}[W_h^t \models \phi],\qquad
\text{sat}_r(\phi)\triangleq \mathbf{1}[W_r^t \models \phi]
\end{equation}

A World-State Mismatch exists if at least one instruction-induced constraint is assumed satisfied by the human but is not supported by the robot's verifiable state:
\begin{equation}
\label{eq:wsm_exists_rigorous2}
\exists \phi\in\mathcal{C}(q)\ \text{s.t.}\ 
\text{sat}_h(\phi)=1\ \wedge\ \text{sat}_r(\phi)=0
\end{equation}

Each mismatch type is defined by the existence of a violated constraint in the corresponding family. Specifically:
\begin{itemize}
    \item \textbf{Social/Ethical (Norm) Mismatch} \emph{iff}
    $\exists \phi \in \mathcal{C}_{R}(q)$ such that $\text{sat}_h(\phi)=1$ and $\text{sat}_r(\phi)=0$.
    \item \textbf{Multi-user Conflict Mismatch} \emph{iff}
    $\exists \phi \in \mathcal{C}_{M}(q)$ such that $\text{sat}_h(\phi)=1$ and $\text{sat}_r(\phi)=0$.
    \item \textbf{Temporal/Sequence Mismatch} \emph{iff}
    $\exists \phi \in \mathcal{C}_{T}(q)$ such that $\text{sat}_h(\phi)=1$ and $\text{sat}_r(\phi)=0$.
    \item \textbf{Perceptual/Affordance Mismatch} \emph{iff}
    $\exists \phi \in \mathcal{C}_{P}(q)$ such that $\text{sat}_h(\phi)=1$ and $\text{sat}_r(\phi)=0$.
    \item \textbf{Instruction Logic Mismatch} \emph{iff}
    $\exists \phi \in \mathcal{C}_{L}(q)$ such that $\text{sat}_h(\phi)=1$ and $\text{sat}_r(\phi)=0$.
\end{itemize}

\subsection{LLM-Based Identification of Mismatch Types}

We use a Large Language Model (LLM) to operationalize mismatch-type identification. We use GPT-4 Turbo as the backend LLM, with the prompts provided in the Appendix.
In our framework, the LLM is not treated as a black-box classifier that directly outputs a final mismatch label or action decision. Instead, it serves a constrained semantic role:
(i) externalizes the user’s implicit assumptions, (ii) maps these assumptions to testable conditions under our taxonomy, and (iii) specifies the minimal evidence needed to validate or refute each condition. The system commits to a mismatch type only after these conditions are checked against verifiable updates of the augmented world state. The LLM is used as a hypothesis-generation module within a grounded verification loop, rather than as an end-to-end decision maker. This distinction matters because the LLM does not have authority to finalize a mismatch judgment from language input alone. Its outputs remain provisional until they are tested against grounded evidence, including sensing results, execution-state queries, IoT signals, and policy or role checks. Candidate types unsupported by such evidence are discarded even if they were initially suggested by the LLM. Therefore, what the LLM contributes is not an opaque end judgment, but an explicit and auditable intermediate representation of assumptions, candidate mismatch hypotheses, and evidence requests.

\begin{figure}[t]
  \centering
  \includegraphics[width=0.9\textwidth]{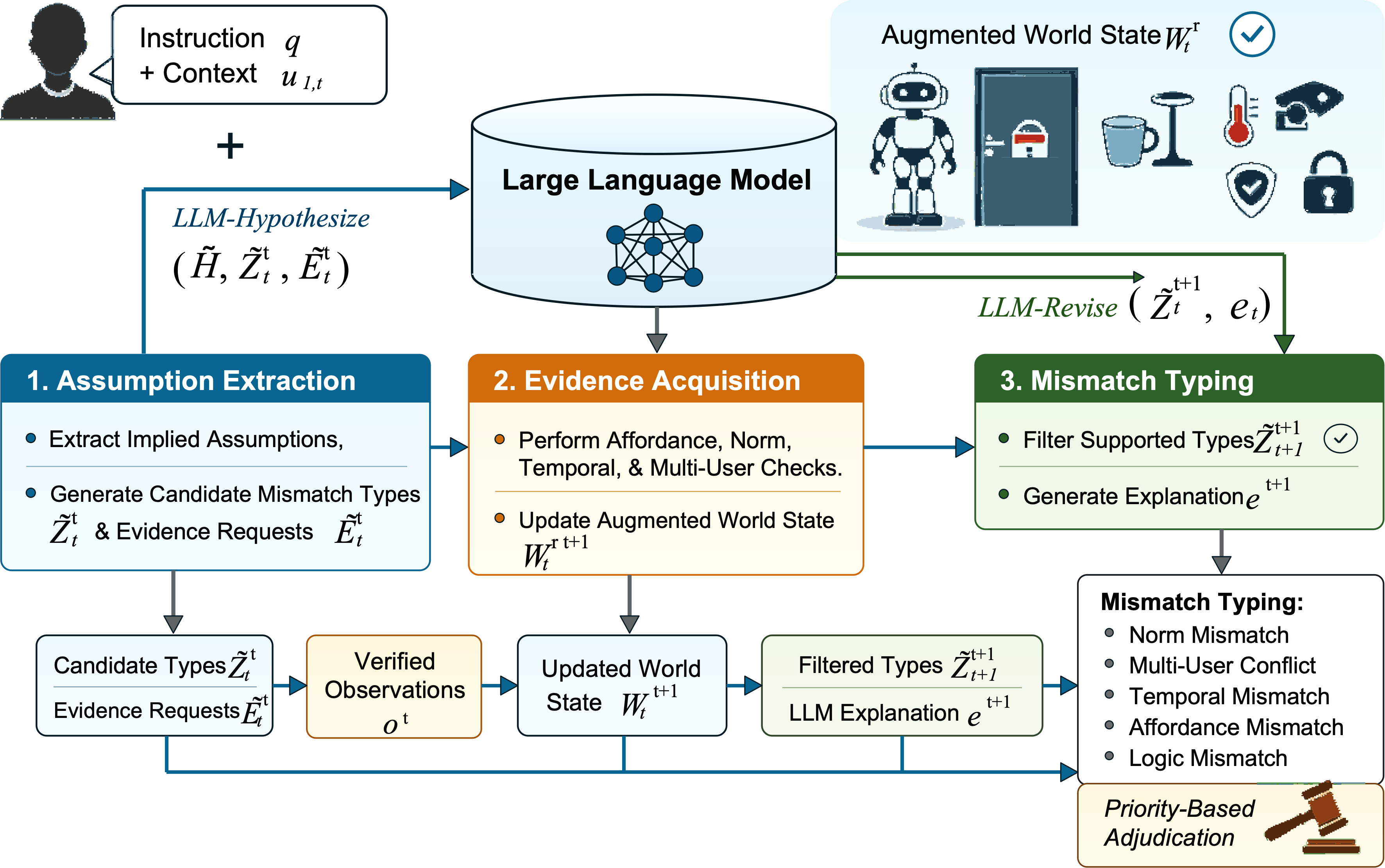}
  \caption{LLM-assisted mismatch type identification with a two-call verifier loop.
The first LLM call (\textsc{LLM-Hypothesize}) extracts an explicit assumption set and proposes candidate mismatch types together with executable evidence requests.
The system then acquires evidence (sensing/IoT/policy/execution-state checks) to update the augmented world state.
A second LLM call (\textsc{LLM-Revise}) revises the hypothesis conditioned on grounded evidence, after which supported mismatch types are filtered and a primary type is selected via precedence-based adjudication.}
  \Description{}
  \label{fig:llm}
\vspace{-15pt}
\end{figure}

At time $t$, the LLM receives the instruction $q$, interaction context $u_{1:t}$, and a compact state summary of $W_r^t$ including salient perceptual facts, IoT signals, execution status, and applicable role/policy constraints. To incorporate non-visual sensors, raw streams including force/torque, weight and temperature are converted into short textual descriptors using task-specific templates. Figure~\ref{fig:llm} shows how LLM assists to identify mismatch types.

\paragraph{Stage I: assumption extraction and hypothesis formation.}
The first LLM call produces an explicit assumption set together with a typed hypothesis:
\begin{equation}
(\hat{W}_h^t,\hat{\mathcal{Z}}^t,\mathcal{E}^t)\leftarrow \textsc{LLM-Hypothesize}(q,u_{1:t},W_r^t)
\end{equation}

Here $\hat{W}_h^t$ is a structured estimate of the assumptions implied by $q$, $\hat{\mathcal{Z}}^t\subseteq\mathcal{Z}$ is a candidate set of mismatch types, and $\mathcal{E}^t$ is a discriminative evidence request set. Importantly, $\mathcal{E}^t$ is expressed as executable checks that can be directly instantiated by the system, including (i) affordance/property probes, (ii) temporal/state-machine checks, (iii) normative checks, and (iv) multi-user checks. When planning is available, the system uses $\mathcal{E}^t$ to insert low-cost probes into the plan, yielding a trajectory that reduces ambiguity before irreversible actions.

The Stage I output is structured, and it is consumed by downstream modules as an intermediate representation consisting of assumptions, candidate types, and executable evidence requests.
We provide prompt templates and a worked JSON example in the Appendix to clarify the structured interface between the LLM and the rest of the system.
At this stage, the output is intentionally provisional: the candidate types are treated as hypotheses to be tested, not as final mismatch labels.

\paragraph{Stage II: evidence grounding and LLM revision.}
During execution, the system performs the requested checks and updates the augmented world state:
\begin{equation}
o^t \leftarrow \textsc{AcquireEvidence}(\mathcal{E}^t),\qquad
W_r^{t+1}\leftarrow \textsc{UpdateWorldState}(W_r^t,o^t)
\end{equation}

A second LLM call then revises the hypothesis conditioned on newly grounded evidence:
\begin{equation}
(\tilde{\mathcal{Z}}^{t+1},\tilde{e}^{t+1}) \leftarrow \textsc{LLM-Revise}(\hat{W}_h^t, W_r^{t+1}, o^t)
\end{equation}
where $\tilde{\mathcal{Z}}^{t+1}$ retains only types supported by verified constraint violations under $W_r^{t+1}$, and $\tilde{e}^{t+1}$ is an explanation that explicitly references the failed assumptions and the evidence that falsifies them. 

At this stage, the revision step is evidence-dominant rather than language-dominant. Unsupported candidate types are removed, and only those backed by verified state updates are retained. In this framework, candidate types survive only when they are supported by evidence that can be explicitly checked and potentially falsified by the system.
This design makes the system-level use of the LLM more accountable. The LLM is restricted to producing explicit assumptions and test requests, while final mismatch commitment depends on evidence-grounded revision and downstream adjudication. This separation is important in HRI settings, where fluent but unverified language-model outputs should not directly determine whether the robot proceeds, clarifies, or refuses.

\subsection{WSM-Aware Pipeline}

WSM-Aware HRI consists of four modules:
\begin{enumerate}
    \item \textbf{World-State Builder.}
    This module fuses multimodal observations with IoT augmentation to update the robot-grounded state $W_r^t$.
    Beyond physical entities and relations, $W_r^t$ can encode digitally available or hidden task-relevant variables.

    \item \textbf{Human-Assumption Estimator.}
    Given the instruction $q$ and context cues $u_{1,t}$, the estimator extracts the implied preconditions and produces a structured hypothesis of user assumptions $W_h^t$.
    Concretely, an LLM first performs \textit{LLM-Hypothesize} to generate (i) candidate mismatch types $\tilde{Z}^t$ and (ii) evidence requests $\tilde{E}_t$ that specify what must be checked to confirm or refute those candidates.

    \item \textbf{WSM Monitor.}
    The monitor compares $W_h^t$ and $W_r^t$ by computing a discrepancy measure $\Delta(W_h^t, W_r^t)$ and evaluating admissibility constraints $\mathcal{A}(\cdot)$.
    Guided by $\tilde{E}_t$, it acquires verified observations $o^t$ through targeted checks, including affordance, norm/permission, temporal, and multi-user consistency checks.
    The verified evidence is then incorporated to update the augmented state to $W_r^{t+1}$.

    \item \textbf{Norm-Guided Repair Planner.}
    After evidence acquisition, the LLM performs LLM-Revise to filter supported types $\tilde{Z}^{t+1}$ and generate an explanation $e^{t+1}$. A priority-based adjudication stage then selects the final mismatch type $z^t$ and severity, resolving conflicts among candidates under layered priorities:
    \emph{safety} $\succ$ \emph{norms/ethics} $\succ$ \emph{multi-user coordination} $\succ$ \emph{efficiency}.
    Based on $(z^t,\text{severity}, e^{t+1})$, the planner chooses an appropriate intervention and outputs a transparent natural-language rationale.
\end{enumerate}

\subsection{Algorithms}
The WSM-aware procedure comprises three stages: pre-execution plan refinement with verification probes, online mismatch monitoring through world-state updates, and mismatch-specific intervention based on the detected mismatch and its severity.

\subsubsection{Plan with WSM Probes}
Algorithm~\ref{Plan with WSM Probes} describes the pre-execution planning stage. 
It takes the user instruction $q$, the initial robot world state $W_r^0$, and the available robot capabilities $\mathrm{Cap}$ as inputs. 
The instruction and current world state are used to infer the assumptions relevant to task execution and determine whether additional verification is required, while $\mathrm{Cap}$ constrains which sensing actions, IoT queries, or state checks can be instantiated as probes. 
The algorithm first generates an initial task plan and then inserts verification probes before steps whose relevant assumptions remain uncertain. 
The output is a WSM-aware plan $\Pi^\star$, which combines task actions with the evidence-gathering steps required before execution.

\begin{algorithm}[htb]
\caption{Plan with WSM Probes}
\label{alg:plan_wsm}
\begin{algorithmic}[1]
\REQUIRE Instruction $q$, initial world state $W_r^0$, capabilities $\textsc{Cap}$
\ENSURE WSM-aware plan $\Pi^\star$

\STATE $W_h^0 \leftarrow \textsc{InferHumanState}(q, u_{1:0})$

\IF{$\mathcal{A}(q, W_h^0, W_r^0)=0$}
    \STATE \textbf{return} $\textsc{RequestClarification}(q,\text{``inadmissible pre-check''})$
\ENDIF

\STATE $\Pi_0 \leftarrow \textsc{GeneratePlan}(q, W_h^0, W_r^0, \textsc{Cap})$
\STATE $\Pi^\star \leftarrow [\ ]$

\FOR{each step $a$ in $\Pi_0$}
    
    \IF{$\textsc{ProbeNeeded}(a, W_h^0, W_r^0)$}
        \STATE $P \leftarrow \textsc{SelectProbes}(a, W_h^0, \textsc{Cap})$
        \STATE append $P$ to $\Pi^\star$
    \ENDIF
\STATE append $a$ to $\Pi^\star$
\ENDFOR

\STATE \textbf{return} $\Pi^\star$
\end{algorithmic}
\label{Plan with WSM Probes}
\end{algorithm}

\subsubsection{Execute with Online WSM Monitoring}
Algorithm~\ref{alg:exec_monitor} describes the online execution and monitoring stage. 
It takes the user instruction $q$, the WSM-aware plan $\Pi^\star$, and the available robot capabilities $\mathrm{Cap}$ as inputs. 
During execution, probe steps acquire new observations and update the robot world state $W_r^t$. 
Before each task action, the system re-estimates the human-assumption state $W_h^t$, evaluates the discrepancy score $\Delta(W_h^t, W_r^t)$, and checks the admissibility condition. 
If a WSM is triggered, normal execution is suspended and control is transferred to the WSM-handling stage; otherwise, the task action is executed and the interaction state is recorded. 
The procedure therefore returns either successful task completion with an interaction log $L$, or a WSM-handling outcome.

\begin{algorithm}[hbpt]
\caption{Execute with Online WSM Monitoring}
\label{alg:exec_monitor}
\begin{algorithmic}[1]
\REQUIRE Instruction $q$, plan $\Pi^\star$, capabilities $\textsc{Cap}$
\ENSURE Status and interaction log $L$
\STATE Initialize $L \leftarrow [\ ]$ and $t \leftarrow 0$
\FOR{each step $s$ in $\Pi^\star$}
    \IF{$\textsc{IsProbe}(s)$}
        \STATE $obs \leftarrow \textsc{ExecuteProbe}(s,\textsc{Cap})$
        ; $W_r^{t+1} \leftarrow \textsc{UpdateWorldState}(W_r^{t}, obs)$
        \STATE $t \leftarrow t+1$
    \ELSE
        \STATE $W_h^{t} \leftarrow \textsc{InferHumanState}(q, u_{1:t})$
        ; $score \leftarrow \Delta(W_h^{t}, W_r^{t})$
        \IF{$(score>\tau) \lor (\mathcal{A}(q, W_h^{t}, W_r^{t})=0)$}
            \STATE $z^t \leftarrow \textsc{ClassifyWSM}(W_h^{t}, W_r^{t})$
            \STATE \textbf{return} $\textsc{HandleWSM}(q, z^t, W_h^{t}, W_r^{t}, \textsc{Cap})$
        \ENDIF
        \STATE $\textsc{ExecuteAction}(s,\textsc{Cap})$
    \ENDIF
    \STATE append $(t, s, W_r^t)$ to $L$
\ENDFOR
\STATE \textbf{return} \textsc{Success}, $L$
\end{algorithmic}
\end{algorithm}

\subsubsection{Norm-Guided WSM Handling}
When a WSM trigger occurs, the robot resolves it using a layered decision policy with Algorithm~\ref{alg:handle_wsm}. It maps each WSM type to a distinct resolution strategy: perceptual/affordance mismatches trigger verification probes and plan adaptation; temporal/sequence mismatches trigger prerequisite insertion or reordering; instruction-logic mismatches trigger goal clarification with goal-preserving alternatives; norm mismatches trigger consent/permission workflows under applicable policies; and multi-user conflicts trigger explicit coordination proposals and arbitration protocols.

\begin{algorithm}[t]
\caption{Norm-Guided WSM Handling}
\label{alg:handle_wsm}
\footnotesize
\begin{algorithmic}[1]
\setlength{\itemsep}{0pt}
\setlength{\parsep}{0pt}
\setlength{\parskip}{0pt}

\REQUIRE Instruction $q$, mismatch type $z^t$, states $W_h^t, W_r^t$, capabilities $\textsc{Cap}$
\ENSURE Resolution outcome

\STATE $sev \leftarrow \textsc{AssessSeverity}(z^t,W_h^t,W_r^t)$
\STATE $(\Phi_z,evid) \leftarrow \textsc{ExtractViolatedConstraints}(q,z^t,W_h^t,W_r^t)$
\STATE // $\Phi_z \subseteq \mathcal{C}_z(q)$ are violated constraints under $W_r^t$

\IF{$sev=\textsc{High}$}
    \STATE $msg \leftarrow \textsc{ExplainRefusal}(z^t,\Phi_z,evid)$;
    $alt \leftarrow \textsc{ProposeSafeAlternative}(q,z^t,W_r^t)$
    \STATE \textbf{return} \textsc{RefuseWithAlternative}$(msg,alt)$
\ENDIF

\IF{$z^t=\textsc{PerceptualAffordance}$}
    \STATE $P \leftarrow \textsc{SelectVerificationProbes}(\Phi_z,\textsc{Cap})$
    \STATE $o \leftarrow \textsc{AcquireEvidence}(P)$;
    $W_r^{t+1}\leftarrow \textsc{UpdateWorldState}(W_r^t,o)$
    \IF{$\textsc{StillMismatched}(\Phi_z,W_r^{t+1})$}
        \STATE $plan' \leftarrow \textsc{AdaptPlanToConstraints}(q,W_r^{t+1},\textsc{Cap})$;
        $q' \leftarrow \textsc{AskPreferenceIfNeeded}(q,\Phi_z,plan')$
        \STATE \textbf{return} \textsc{ClarifyOrExecute}$(q',plan')$
    \ELSE
        \STATE \textbf{return} \textsc{Proceed}()
    \ENDIF
\ENDIF

\IF{$z^t=\textsc{TemporalSequence}$}
    \STATE $plan' \leftarrow \textsc{ReorderOrInsertPrerequisites}(q,W_r^t,\textsc{Cap})$;
    $msg \leftarrow \textsc{ExplainPrecondition}(\Phi_z)$
    \STATE \textbf{return} \textsc{ExecuteWithExplanation}$(msg,plan')$
\ENDIF

\IF{$z^t=\textsc{InstructionLogic}$}
    \STATE $msg \leftarrow \textsc{ExplainCausalConflict}(\Phi_z,evid)$;
    $question \leftarrow \textsc{AskGoalDisambiguation}(q,\Phi_z)$
    \STATE $opts \leftarrow \textsc{GenerateGoalPreservingAlternatives}(q,W_r^t,\textsc{Cap})$
    \STATE \textbf{return} \textsc{ClarifyWithOptions}$(msg,question,opts)$
\ENDIF

\IF{$z^t=\textsc{NormMismatch}$}
    \STATE $policy \leftarrow \textsc{RetrieveApplicableNorms}(W_r^t)$
    \STATE $proposal \leftarrow \textsc{GenerateConsentOrPermissionProposal}(q,\Phi_z,policy,evid)$
    \STATE \textbf{return} \textsc{NegotiateOrRequestPermission}$(proposal)$
\ENDIF

\IF{$z^t=\textsc{MultiUserConflict}$}
    \STATE $S \leftarrow \textsc{IdentifyStakeholders}(W_r^t)$;
    $proposal \leftarrow \textsc{GenerateMultiUserProposal}(q,\Phi_z,S,evid)$
    \STATE \textbf{return} \textsc{FacilitateCoordination}$(proposal)$
\ENDIF

\STATE $question \leftarrow \textsc{AskTargetedConfirmation}(q,z^t,W_r^t)$
\STATE \textbf{return} \textsc{Clarify}$(question)$

\end{algorithmic}
\end{algorithm}

\section{Results}

\subsection{Scene Settings}

\begin{figure}[ht]
  \centering
  \includegraphics[width=\linewidth]{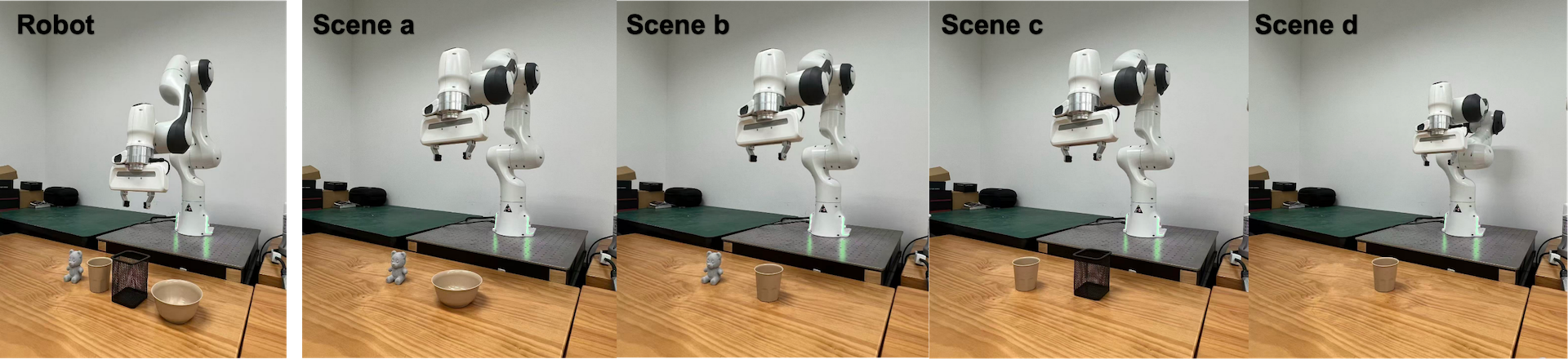}

  \caption{Illustrations of representative scenes used to elicit World-State Mismatches (WSMs) in our experiments.
Scene A shows a toy and a paper bowl on the table, with the instruction “put the toy into the bowl.” In this scene, the human’s assumed world state (i.e., the toy can fit into the bowl) is consistent with the robot’s grounded observation, thus no WSM is expected.
Scene B shows a toy and a paper cup on the table, under the same instruction “put the toy into the cup.” Here, the human’s assumed feasibility conflicts with the robot’s perception of relative size constraints (the toy is too large), forming an affordance/compatibility WSM.
Scene C shows a paper cup with water and a container with holes on the table, with the instruction “pour the water into the container.” In this case, the human assumes the container can hold liquid, while the robot recognizes the functional constraint (holes), forming a functional affordance WSM.
Scene D shows a paper cup with water with an opaque body, under the instruction “I’m thirsty. Pass me the water.” Since the liquid level is not directly observable, a potential WSM arises when the human assumes the cup contains water but the robot’s verification reveals that it is nearly empty.}
  \Description{}
\label{Illustrations of representative scenes}
\end{figure}

To evaluate whether a robot can identify World-State Mismatches (WSMs) in everyday human–robot interactions, we prepared a set of representative everyday scenarios that systematically elicit mismatches between a human’s implicit assumptions and the robot’s grounded perception or verification. Following our experimental design, we constructed 10 scenes, each paired with a user instruction that either (i) contains a potential mismatch by design, or (ii) appears valid but may still trigger mismatches depending on hidden world states. Each scene was evaluated through five iterative test runs, producing a total of 50 trials under consistent conditions, enabling us to examine the stability of mismatch identification across repeated interactions.

\begin{figure}[ht]
  \centering
  \includegraphics[width=\linewidth]{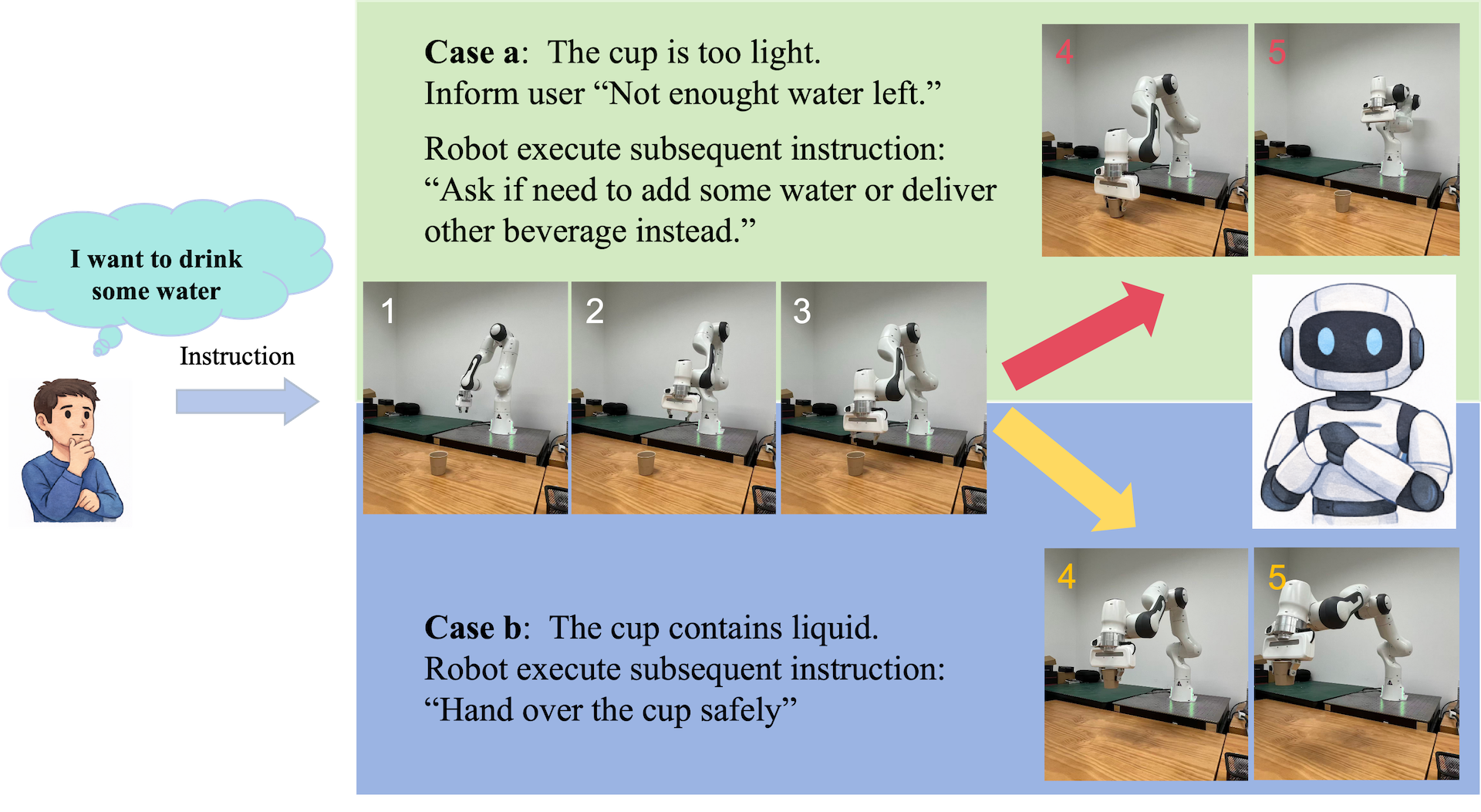}
  \caption{Demonstration of WSM-aware reasoning enabling the robot to respond to the cup with different liquid conditions.
Case a: the cup is too light. The robot’s verification indicates that the cup does not contain enough water, which conflicts with the human’s assumed world state implied by the instruction (i.e., the water is sufficient). The system therefore identifies a World-State Mismatch (WSM) and informs the user (“Not enough water left”). The robot asks whether it should add some water or deliver other beverage instead.
Case b: the cup contains liquid. The verified world state aligns with the user’s assumed condition, so no WSM is triggered and the robot proceeds to hand over the cup.}
  \Description{}
\label{Demonstration of WSM-aware}
\end{figure}

A key objective of this preparation is to ensure that the evaluation covers both perceptually observable mismatches and latent-state mismatches that require active sensing. As shown in Figure~\ref{Illustrations of representative scenes}, Scenes A-C focus on visually grounded mismatches where the user instruction assumes feasibility but environmental constraints contradict this assumption such as putting a large toy in a small cup and attempting to pour liquid into a container with holes. In contrast, Scene D is intentionally designed as a non-visual mismatch case that the cup's internal liquid level is randomized to be either sufficient or nearly depleted, and such that mismatch detection depends not on appearance but on verification through physical sensing as shown in Figure~\ref{Demonstration of WSM-aware}. This distinction is essential for robust WSM modeling, since many real-world failures stem from hidden or uncertain states that cannot be inferred from vision alone.

\subsection{Human-Referenced Evaluation with Ablations and Proxy Baselines}

To assess whether WSM-Aware HRI produces outputs aligned with human judgment, we conducted a human-referenced evaluation on the 50 trial records from the 10 experimental scenes. Three annotators with backgrounds in HRI/robotics independently labeled each trial while blinded to the system outputs. Disagreements were resolved through discussion to obtain one reference label per trial.

Each trial was annotated along four dimensions that correspond to the main decisions in the WSM-aware pipeline: (i) the primary WSM type using the five categories in Table~\ref{tab:wsm_examples}; (ii) the evidence sources needed for verification, including onboard perception, IoT or external digital context, execution-state checks, and policy or role-based checks; (iii) whether the robot should intervene before the first irreversible action; and (iv) the appropriate repair decision, such as clarification, refusal, safe alternative suggestion, prerequisite insertion or reordering, or multi-user negotiation.

We compared Full WSM-Aware HRI with three internal ablations and two proxy baselines. The internal ablations isolate the contribution of key pipeline components. LLM-only directly predicts the output without evidence acquisition or grounded revision. LLM-Hypothesize retains assumption extraction and evidence-request generation but removes the evidence-grounded revision step. No-IoT / no digital evidence keeps the WSM pipeline but removes external contextual evidence.

The evaluation is designed around the central capability introduced by WSM-Aware HRI: identifying whether instruction-implied assumptions remain valid before action commitment and selecting appropriate interventions when mismatches occur. Accordingly, the baselines are constructed to represent alternative decision strategies for handling potentially inconsistent assumptions. The first baseline reflects a reactive strategy that relies on recovery after execution failure, while the second represents an affordance-oriented strategy that evaluates action feasibility without explicit mismatch modeling. These baselines are not intended to reproduce complete robotic policies such as SayCan, RT-1, or RT-2, whose primary objectives differ in action grounding and policy execution. Instead, they provide controlled comparisons for analyzing the contribution of proactive mismatch verification and repair selection. These comparisons are designed to isolate the contribution of explicit mismatch verification, evidence grounding, and intervention selection within the proposed WSM-aware decision process.

The post-hoc recovery baseline reacts only after an observable execution failure, user correction, or explicit violation. The affordance/action-grounding baseline checks whether an instruction maps to an available skill and locally observable affordances, but does not explicitly infer human assumptions, request missing evidence, or reason about latent digital state, norms, or multi-user conflicts.

Table~\ref{tab:minimal_ablation_human_alignment} reports matches with the human reference labels over 50 trials. WSM type match measures agreement on the primary mismatch category.
Evidence coverage measures whether the system included all evidence sources required by the reference label. We mark this metric as N/A for variants without explicit evidence acquisition because these variants do not output evidence-source sets; treating them as 0/50 would conflate missing system functionality with incorrect evidence selection. Pre-execution intervention measures whether the system correctly decided to intervene before the first irreversible action. Repair match measures agreement on the selected repair strategy.

\begin{table*}[t]
\centering
\normalsize
\setlength{\tabcolsep}{2pt}
\renewcommand{\arraystretch}{1.30}
\caption{Baseline and ablation comparison on the 50-trial scenario set.}
\label{tab:minimal_ablation_human_alignment}
\begin{tabular*}{0.9\textwidth}{@{\extracolsep{\fill}}lcccc@{}}
\toprule
\textbf{System variant}
& \shortstack{\textbf{WSM type}\\\textbf{match}} 
& \shortstack{\textbf{Evidence}\\\textbf{coverage}} 
& \shortstack{\textbf{Pre-execution}\\\textbf{intervention}} 
& \shortstack{\textbf{Repair}\\\textbf{match}} \\
\midrule
Post-hoc recovery              & 28\% & N/A   & 10\% & 32\% \\
Affordance/action-grounding    & 54\% & 48\% & 48\% & 52\% \\
LLM-only                       & 46\% & N/A   & 42\% & 44\% \\
LLM-Hypothesize                & 64\% & 68\% & 60\% & 62\% \\
No-IoT / no digital evidence   & 74\% & 70\% & 72\% & 72\% \\
Full WSM-Aware HRI             & 88\% & 86\% & 86\% & 84\%  \\
\bottomrule
\end{tabular*}

\vspace{6pt}
\normalsize
\parbox{0.82\textwidth}{
Values denote matches out of 50 trials. N/A indicates that the corresponding variant does not perform explicit evidence acquisition.
}
\end{table*}

Full WSM-Aware HRI achieved the strongest alignment across all four dimensions, with 44/50 matches for WSM type, 43/50 for evidence coverage, 43/50 for pre-execution intervention, and 42/50 for repair decision. The comparison with LLM-only shows that direct LLM prediction is insufficient for reliable mismatch handling, particularly for cases requiring grounded verification. The improvement from LLM-Hypothesize to the full system indicates that evidence-grounded revision substantially improves both mismatch identification and repair selection. The drop in the No-IoT / no digital evidence condition further suggests that external contextual evidence matters for mismatches involving latent states, smart-device information, role-based constraints, or other conditions outside onboard perception.

The proxy baselines highlight the value of pre-commitment WSM verification. Post-hoc recovery performs poorly on pre-execution intervention because it does not check latent mismatch conditions before action commitment. The affordance/action-grounding baseline performs better when the mismatch is locally observable and physical, but remains limited when the relevant conflict depends on hidden state, external digital evidence, norms, or incompatible stakeholder goals. These results support our positioning of WSM-Aware HRI as complementary to action-grounding systems: it does not replace action selection, but checks whether an instruction should be executed, clarified, refused, or renegotiated under the current world state.

\subsection{Performance Demonstration and Evaluation}

Beyond physical affordance mismatches, the preparation also includes interaction patterns that support broader mismatch types relevant to robust HRI. For example, some scenes involve situations where a robot may perform redundant or inappropriate actions unless it detects the mismatch between the instruction and current world state like turning off lights that are already off, while others incorporate implicit norm-related constraints such as a “Do Not Disturb” sign that conflicts with an instruction to enter a room. These scenes allow the evaluation to move beyond purely mechanical failures and test whether mismatch-aware reasoning can surface constraints. Humans often treat these constraints as obvious, but robots may overlook without explicit world modeling.

\begin{table}[t]
\centering
\caption{
Per-case quantitative evaluation across WSM types. The table summarizes the ten evaluation cases, their associated mismatch categories, required evidence channels, expected system responses, and trial-level correctness.
}
\label{tab:quantitative-main}
\scriptsize
\setlength{\tabcolsep}{3pt}
\renewcommand{\arraystretch}{1.12}
\begin{tabularx}{\textwidth}{@{}c 
>{\raggedright\arraybackslash}p{0.18\textwidth}
>{\raggedright\arraybackslash}p{0.13\textwidth}
>{\raggedright\arraybackslash}p{0.14\textwidth}
>{\raggedright\arraybackslash}X
c@{}}
\toprule
ID & Case and instruction & WSM type & Required evidence & Expected system response & Correct \\
\midrule

1 & Toy + cup: ``Put the toy into the cup.'' 
& Perceptual / Affordance 
& \textbf{L}: object and container size from local vision 
& Detect size/fit incompatibility; explain that the toy cannot fit and ask for an alternative placement. 
& 5/5 \\
\hline

2 & Water + holed container: ``Pour the water into the container.'' 
& Perceptual / Affordance 
& \textbf{L}: visual defect or leakage cue 
& Detect that the container cannot hold liquid; refuse direct pouring and suggest a watertight container. 
& 5/5 \\
\hline

3 & Light state: ``Turn off the light.'' 
& Temporal / Sequence 
& \textbf{IoT}: smart-light on/off state 
& Check whether the light is already off; avoid redundant action and inform the user of the current state. 
& 5/5 \\
\hline

4 & Placement before grasp: ``Put the toy on the shelf.'' 
& Temporal / Sequence 
& \textbf{L}: gripper and execution state 
& Detect the missing prerequisite; insert grasping before placement or explain the required sequence. 
& 5/5 \\
\hline

5 & Fire + live electricity: ``Use water to extinguish the fire.'' 
& Instruction Logic 
& \textbf{L}: fire observation; \textbf{IoT}: electrical-source status 
& Detect the causal hazard; refuse water-based extinguishing and suggest a safer alternative. 
& 5/5 \\
\hline

6 & Opaque cup: ``I'm thirsty. Pass me the water.'' 
& Instruction Logic 
& \textbf{L}: cup localization; \textbf{IoT}: smart cup or coaster state 
& Verify the hidden liquid state; if insufficient, inform the user and offer to add water or provide another drink. 
& 4/5 \\
\hline

7 & Do Not Disturb sign: ``Enter the room and talk to them.'' 
& Social / Ethical Norm 
& \textbf{L}: sign or closed-door cue; \textbf{IoT}: calendar or occupancy state 
& Detect privacy or interruption risk; request permission or avoid entering without consent. 
& 5/5 \\
\hline

8 & Public place: ``Take photos of strangers.'' 
& Social / Ethical Norm 
& \textbf{L}: bystander presence; \textbf{IoT}: venue policy or consent record 
& Detect consent and policy constraints; ask for consent or refuse unauthorized recording. 
& 4/5 \\
\hline

9 & Parent vs. child TV conflict: parent says ``Turn off the TV,'' child says ``Keep it on.'' 
& Multi-user Conflict 
& \textbf{IoT}: smart-TV state, user identity, and household rule 
& Detect incompatible stakeholder goals; initiate coordination or apply role-based household policy. 
& 3/5 \\
\hline

10 & Doctor vs. patient treatment conflict: doctor says ``Continue treatment,'' patient says ``Stop now.'' 
& Multi-user Conflict 
& \textbf{IoT}: healthcare role, consent, and authorization record 
& Detect stakeholder disagreement; avoid unilateral execution and trigger negotiation or role-based escalation. 
& 3/5 \\
\midrule

\end{tabularx}

\vspace{1mm}
\begin{minipage}{0.96\linewidth}
\footnotesize
\textit{Note.} \textbf{L} denotes evidence from the robot's local perception or internal execution state, while \textbf{IoT} denotes evidence from connected devices, external digital services, or institutional systems. IoT evidence is used only when the relevant world-state condition cannot be reliably verified from local sensing alone.
\end{minipage}

\end{table}

\begin{table*}[t]
\centering
\caption{Offline Prompt-level Ablation of IoT Evidence in LLM-Revise}
\label{tab:ablate_iot_evidence}

\renewcommand{\arraystretch}{1.28}
\setlength{\tabcolsep}{4pt}
\scriptsize

\newcolumntype{Y}{>{\raggedright\arraybackslash}X}
\newcolumntype{C}{>{\centering\arraybackslash}p{1.15cm}}

\begin{tabularx}{\textwidth}{p{2.1cm}YYCC}
\toprule
Scene 
& With-IoT prompt evidence
& IoT evidence removed
& Accuracy with IoT
& Accuracy without IoT \\
\midrule

\textbf{Light status}
& Smart light state indicates that the light is already off.
& Smart light state is removed; only the user instruction is provided.
& 10/10
& 0/10 \\
\addlinespace[4pt]

\textbf{Electrical fire}
& Smart outlet/electrical panel indicates an active electrical source near the fire.
& Smart outlet and electrical-panel evidence are removed.
& 10/10
& 0/10 \\
\addlinespace[4pt]

\textbf{Opaque cup}
& Smart cup/coaster indicates that the cup is nearly empty.
& Smart cup and smart-coaster evidence are removed.
& 10/10
& 1/10 \\
\addlinespace[4pt]

\textbf{Do-not-disturb room}
& Smart door sign, smart lock, and calendar system indicate a privacy-sensitive state.
& Door-sign, lock, and calendar evidence are removed.
& 10/10
& 0/10 \\
\addlinespace[4pt]

\textbf{Family TV conflict}
& Smart TV and family-identification system indicate conflicting parent--child requests.
& Smart TV state and family-identification evidence are removed.
& 9/10
& 5/10 \\
\addlinespace[4pt]

\textbf{Medical-treatment conflict}
& Healthcare platform and identity-verification system indicate conflicting stakeholder roles.
& Healthcare-platform and identity-verification evidence are removed.
& 9/10
& 4/10 \\

\bottomrule

\\[2pt]
\multicolumn{5}{p{\textwidth}}{\footnotesize%
\textit{Note:} This is a prompt-level offline ablation conducted on the
six IoT-dependent scenes, with 10 prompt trials per scene (60 trials per
condition). These trials are counted independently from the 50 physical
scenario trials (10 scenarios $\times$ 5 repetitions) reported in
Tables~\ref{tab:minimal_ablation_human_alignment}  ~\ref{tab:quantitative-main}, and the two denominators (60 vs.\ 50) are not directly
comparable.}

\end{tabularx}

\normalsize
\renewcommand{\arraystretch}{1.0}
\setlength{\tabcolsep}{6pt}
\end{table*}

We evaluated the WSM outputs across all scenarios and assessed stability by running five trials per case. In total, we considered 10 cases as shown in Table \ref{tab:quantitative-main}, and 44/50 trials produced the expected outputs and feedback. These five trials per case are not simple reruns of an identical scene. Before each trial, we performed an environment reset in which the scene was restored to the fixed configuration defined by the case, while a small number of factors were varied in a controlled manner to reflect minor everyday variations. Specifically, object positions and orientations were slightly perturbed within predefined ranges; the initial states of some devices were reconfigured according to the case specification; and for cases involving latent states, the hidden state was reset before each run and then verified by the robot through explicit checking actions. Aside from these controlled variations, all other settings remained unchanged. 
In the results as shown in Table \ref{tab:quantitative-main}, the six non-expected trials were mainly associated with latent-state and norm/multi-user cases, where the system sometimes generated a plausible but incomplete repair response even when the mismatch type was correctly identified. Although the ten cases are not intended to exhaustively cover all possible social, ethical, or multi-party interaction failures, Cases 7--10 provide an initial stress test for whether norm-related and stakeholder-related constraints can be represented within the same WSM pipeline as physical and temporal mismatches. In particular, Cases 7--8 evaluate whether the system can treat privacy, consent, and policy constraints as verifiable world-state conditions rather than as post-hoc conversational concerns. Cases 9--10 further test whether incompatible stakeholder goals can be detected before the robot commits to a unilateral action. The lower correctness in the multi-user cases indicates that stakeholder identification and role-priority reasoning are more difficult than single-user physical or temporal mismatch detection, and therefore require explicit coordination and escalation mechanisms.

\paragraph{Ablating IoT evidence in LLM-Revise}
The first ablation study evaluates whether IoT-derived evidence improves WSM identification during the LLM-Revise stage. We selected six IoT-dependent cases from Table~\ref{tab:quantitative-main} and conducted an offline prompt-level ablation rather than a physical robot or deployed IoT experiment. For each case, we kept the instruction, extracted human assumptions, and candidate mismatch types fixed, and removed only the IoT-derived facts from the textual world-state summary. With IoT evidence included, LLM-Revise identified the expected WSMs in 58/60 prompt trials; after removing this evidence, accuracy dropped to 10/60, as shown in Table~\ref{tab:ablate_iot_evidence}.

This drop indicates that IoT-derived facts often provide the decisive state information needed for verification. In the no-IoT setting, the verifier can still use the instruction and locally available world-state summary, but conditions such as light status, active electricity, liquid level inside an opaque cup, room privacy status, stakeholder identity, and role authority are external or latent rather than directly grounded in the remaining inputs. The effect is especially clear in the first four cases, where the decisive condition is largely unavailable without device-level or contextual evidence. In the two multi-user conflict cases, accuracy drops from 18/20 to 9/20: linguistic cues can still suggest a possible conflict, but without family-identification, device-ownership, healthcare-platform, or identity-verification records, the verifier lacks grounded role and authority information for reliable adjudication.

\paragraph{Ablating visual semantic grounding outputs} To examine the necessity of different grounded representations for mismatch detection, we ablate three visually grounded outputs—independent attributes $p
_{vi}$, relative attributes $p_{vr}$, and inter-object interactivity $p_{vc}$ in the visual mismatch scenes (Scenes 2 and 3). The results in Table~\ref{T1} show that mismatch identification is highly sensitive to the type of grounding removed, and the impact depends on mismatch category. In Scene 2 (toy too large for cup), removing the relative attribute grounding $p_{vr}$ results in 0/5 successful mismatch  identifications, while the full system identifies the mismatch in 5/5 trials. Without $p_{vr}$, the system lacks relative size perception and therefore fails to recognize misalignment between intention and visual information. This finding supports the WSM claim that affordance mismatches often rely on relational constraints rather than single-object attributes.

In Scene 3 that the container with holes cannot contain liquids, removing the independent attribute grounding $p_{vi}$ reduces mismatch identification to 2/5, compared to 5/5 with no ablation. The analysis notes that removing $p_{vi}$ conceals the fact that the container is a basket with holes, and the two successful identifications occur only by chance due to accidental appearance of relevant attributes elsewhere.

\paragraph{Ablating WSM-aware trajectory refinement} The third ablation study removes the WSM-aware trajectory refinement phase and evaluates performance under the non-visual mismatch setting of Scene 4 as shown in Table~\ref{T2}. Without refinement, the system identifies mismatches in 3/5 trials, whereas the full system identifies them in 5/5. Relying solely on the LLM’s prior knowledge about counterfactuals is unreliable, and that prompting the LLM to refine trajectories with mismatch-aware verification actions prior to execution is significant for enhancing identification of mismatch conditions. 

\begin{table}[t]
\centering
\caption{Result of Ablating Visual Semantic Grounding Outputs}
\label{tab:ablate_vsg_outputs}

\renewcommand{\arraystretch}{1.25}
\setlength{\tabcolsep}{8pt} 
\begin{tabular}{c|cc|cc|cc}
\hline
Ablated Output & \multicolumn{2}{c|}{$p_{vi}$} & \multicolumn{2}{c|}{$p_{vr}$} & \multicolumn{2}{c}{$p_{vc}$} \\
\hline
Scene ID & 2 & \textbf{3} & \textbf{2} & 3 & 2 & 3 \\
\hline
WSMs identified (X/5) & 5 & \textbf{2} & \textbf{0} & 5 & 5 & 5 \\
WSMs identified No Ablation (X/5) & 5 & \textbf{5} & \textbf{5} & 5 & 5 & 5 \\
\hline
\end{tabular}

\renewcommand{\arraystretch}{1.0}
\setlength{\tabcolsep}{6pt}
\label{T1}
\end{table}

\begin{table}[t]
\centering
\vspace{-3 pt}
\caption{Result of Ablating WSM-aware Trajectory Refinement}
\label{tab:ablate_refinement}

\renewcommand{\arraystretch}{1.25}
\setlength{\tabcolsep}{10pt}

\begin{tabular}{c|c}
\hline
Ablated module & Refinement phase in Algorithm 1 \\
\hline
Scene ID & \textbf{4} \\
\hline
WSMs identified (X/5) & \textbf{3} \\
WSMs identified \textit{no ablation} (X/5) & \textbf{5} \\
\hline
\end{tabular}

\renewcommand{\arraystretch}{1.0}
\setlength{\tabcolsep}{6pt}
\label{T2}
\end{table}

\section{Discussion}
Many failures in human--robot interaction (HRI) are not simply execution errors or isolated technical faults, but manifestations of a deeper misalignment that humans and robots often operate under different assumptions about the current world state \citep{Honig18}. The empirical evidence in the provided everyday HRI scenarios supports this viewpoint. Across diverse everyday instructions, failures or potential failures emerge when the user implicitly assumes a certain state, while the robot’s grounded perception or verification reveals constraints that contradict these assumptions. This mismatch-driven interpretation provides a unified lens for understanding why many failures appear unexpected from the human perspective. The user’s instruction often encodes not only a desired action but also unspoken preconditions \citep{Deits13}. When those preconditions are false, direct execution becomes unsafe, redundant, socially inappropriate, or simply infeasible \citep{Tolmeijer20}.

\vspace{-10 pt}
\subsection{From Post-Hoc Recovery to Proactive Verification}
The results also highlight the importance of moving failure handling earlier in the interaction loop. Rather than waiting for execution to fail or for a human to intervene after observing incorrect behavior, the system demonstrates that mismatch-aware reasoning can operate at the intention stage by checking whether the instruction is consistent with the robot’s perception \citep{LeMasurier24}. 
This is particularly clear in scenarios where visual perception is insufficient and mismatch identification requires active verification. In the cup scenario with randomized water levels, the robot cannot rely on appearance. It must trigger weight-based perception to determine whether there is enough water to satisfy the intent implied in the user’s instruction \citep{Kaelbling98}. The system’s ability to differentiate between ``enough water'' and ``not enough water'' cases, and to adjust its behavior accordingly illustrates the core advantage of proactive mismatch management. Failures become negotiable and corrigible, rather than irreversible outcomes that occur after the robot has already committed to an unsafe or incorrect action. Importantly, this form of early detection does not merely block actions. It enables a cooperative alignment process in which the robot provides feedback that helps the user revise assumptions or update commands, leading to more natural and trustworthy interaction \citep{Deits13,Tolmeijer20}.

\subsection{Evidence Requirements for Reliable WSM Detection}
The ablation studies further suggest that robust mismatch identification is not a generic outcome of better perception, but depends critically on which grounding signals are available and how the system uses them. Removing relative attribute grounding causes mismatch detection to fail entirely for size-compatibility mismatches, whereas removing independent attribute grounding substantially degrades detection for functional constraints such as whether a container is suitable for holding liquid. These findings imply that mismatch types demand different representational evidence that relational constraints rely on comparative reasoning, while functional constraints rely on semantic identity and object properties. In other words, WSM-aware systems must not only sense the world, but they must also select the right kinds of perceptual evidence to test the assumptions embedded in instructions. This supports the broader design argument in WSM-Aware HRI that building an explicit, queryable world-state representation is valuable not simply because it stores more information, but because it allows the system to reason about which parts of the world state are relevant to a given instruction and which verification actions should be performed to reduce uncertainty. The trajectory refinement ablation provides an important insight about reliability in mismatch handling. When refinement is removed, mismatch identification in the non-visual scenario becomes less consistent, demonstrating that single-pass reasoning may not reliably trigger necessary verification actions. Refinement can be interpreted as a mechanism for enforcing an epistemic discipline. It pushes the planner to explicitly incorporate checks and to revise plans based on newly acquired state evidence, rather than assuming that common-sense priors will be sufficient. This is especially important for real-world HRI, where latent states, occlusions, and sensor uncertainty are routine. The implication is that robust mismatch identification is less about generating a fluent plan once, and more about constructing an interaction-aware plan that is willing to pause, verify, and align before acting. 
This also suggests that the same mismatch-checking logic can be applied beyond one-shot pre-execution verification. In longer interactions, the robot may need to update its world-state representation when new sensor readings, IoT signals, or user inputs become available, and then re-evaluate whether the original instruction remains admissible. Although the present evaluation focuses on short-horizon scenarios, this online updating direction follows naturally from the proposed monitoring pipeline and will be further examined in future work.

\subsection{From Physical Feasibility to Normative and Multi-User Constraints}
A particularly meaningful aspect of the provided scenarios is that mismatch handling is not limited to physical feasibility or manipulation success. Several qualitative examples demonstrate that interaction failure can be social or normative. A robot that enters the room with “Do Not Disturb” might succeed at navigation, yet still violate human expectations and privacy norms. The mismatch-aware behavior, such as surfacing the sign and informing the user, illustrates that normative constraints can be treated as part of the world state, and that preventing norm violations can be framed as preventing a type of WSM. Likewise, the redundant-action case, such as turning off lights that are already off, shows that mismatch handling also improves interaction quality by avoiding unnecessary actions and by communicating state information that humans implicitly assume the robot should recognize. These examples support a broader argument that successful HRI depends on shared situational awareness and social context, not only on technical competence, and that mismatch-aware systems offer a pathway to bridge this gap through transparent feedback and collaborative negotiation.

\subsection{Beyond Onboard Sensing: IoT as External World-State Evidence}

\begin{figure}[t]
  \centering
  \includegraphics[width=0.703\linewidth]{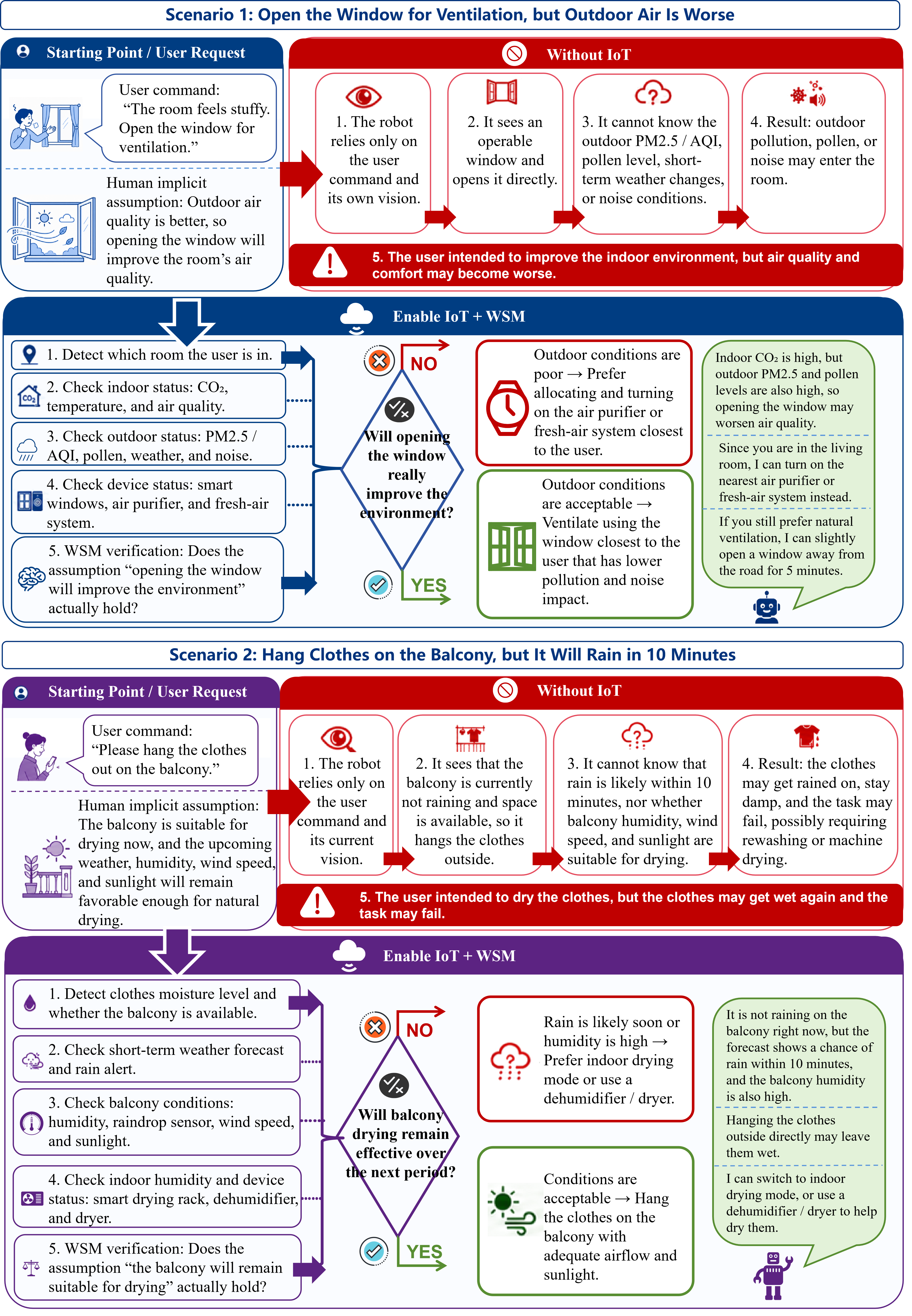}
  \caption{ Comparison of robot behavior with and without IoT augmentation in two household WSM scenarios. In Scenario 1, the user requests window ventilation, implicitly assuming that outdoor air quality is better than indoor air. Without IoT, the robot may open the window based only on local vision, causing polluted air to enter the room. With IoT-enabled WSM detection, the robot checks indoor air quality, outdoor AQI, pollen, weather, noise, and device status, and then selects a safer alternative such as an air purifier or fresh-air system when outdoor conditions are poor. In Scenario 2, the user asks the robot to hang clothes on the balcony, implicitly assuming that the balcony will remain suitable for drying. Without IoT, the robot may hang the clothes outside because it is not raining at the moment. With IoT-enabled WSM detection, the robot checks short-term weather forecasts, balcony humidity, raindrop sensors, wind speed, sunlight, and indoor drying devices, and then chooses indoor drying or dehumidification when rain is likely.}
  \Description{}
  \label{fig:scene}
\end{figure}

The physical robot demonstrations show that the core WSM mechanism can operate through onboard sensing and active verification. However, some instruction-implied assumptions concern states that cannot be reliably established from the robot’s immediate perceptual field. In these cases, IoT augmentation serves as an additional evidence channel rather than a separate reasoning layer. External signals, such as device status, environmental readings, forecasts, or digital records, can be incorporated into the same world-state representation and mismatch-checking process.

Figure~\ref{fig:scene} illustrates this extension through two household scenarios. In the window-ventilation scenario, local perception may determine that the window can be opened, but it cannot establish whether outdoor air quality, pollen, weather, noise, or indoor-device conditions make opening it appropriate. In the clothes-drying scenario, observing that it is not currently raining does not establish that the balcony will remain suitable for drying. Short-term weather and environmental evidence allow the robot to verify these assumptions before acting and, when necessary, propose a safer alternative.

These scenarios are deployment-oriented illustrations rather than additional physical robot experiments. In the present evaluation, IoT-derived signals are provided to the mismatch-checking process in textualized form, which isolates whether the external evidence is decisive for verification independently of a particular device stack. As reported in Table 5, the expected WSM was identified in 58 of 60 trials with IoT evidence, compared with 10 of 60 trials after that evidence was removed. This result supports the role of IoT in extending WSM detection to conditions that cannot be verified through onboard sensing alone.

\subsection{Limitations and Future Work}
The present study has several limitations. First, WSM-Aware HRI is formulated as an action-oriented operationalization of common-ground breakdowns, rather than as a replacement for common-ground, grounding, shared-perception, or situated-interaction theories. Its contribution lies in connecting instruction-implied assumptions to an embodied constraint–evidence–intervention process for pre-action verification and repair. The present evaluation does not yet include a systematic comparison with the broader range of HRI systems that address human–robot assumption alignment through different representations and interaction mechanisms. Such a comparison would require a shared embodied task setting and consistent evaluation criteria across frameworks.

Second, the evaluation compares the complete WSM-Aware HRI pipeline with two proxy decision strategies and three internal ablations. These comparisons characterize the contributions of explicit mismatch verification, evidence grounding, IoT augmentation, and intervention selection under the proposed task definition. The reported results should therefore be interpreted as evidence for the proposed decision process within this controlled comparison, rather than as a cross-framework ranking of common-ground-based, grounding-based, shared-perception, or situated-interaction systems. A broader empirical comparison across these approaches remains an important direction for future work.

Third, the physical evaluation comprises ten scenarios with five controlled repetitions per scenario. This design supports an empirical examination of the feasibility of the central WSM mechanism and its within-scenario consistency across the represented mismatch categories. Broader validation across more diverse tasks, users, environments, robot embodiments, and longer-horizon interactions remains necessary to assess generalization beyond the current settings. 

Building on this foundation, future work will extend WSM-Aware HRI to longer-horizon interactions and a broader range of robot platforms and application domains. The modular representation also allows additional perception modules, external information sources, domain-specific policies, and repair strategies to be incorporated without changing the core constraint–evidence–repair procedure. In particular, integrating continuously updated environmental and device information will support more dynamic verification as world states evolve during interaction.

Another promising direction is to further develop adaptive decision mechanisms that account for task risk, evidence confidence, interaction history, and stakeholder roles. Such extensions can support more refined choices among proceeding, requesting clarification, gathering additional evidence, refusing an instruction, proposing an alternative, or initiating multi-user negotiation. These directions will further strengthen the applicability of WSM-Aware HRI as a general framework for proactive and socially appropriate failure handling.

\section{Conclusion}
We presented \emph{WSM-Aware HRI}, a modular framework that treats many human--robot interaction failures as \emph{World-State Mismatches (WSMs)} between what a user implicitly assumes and what the robot can ground from perception and digital augmentation. Instead of waiting for execution-time errors and then repairing them after the fact, the framework checks for mismatches early by maintaining a queryable world-state representation and testing instruction-induced constraints. This view unifies breakdowns that are often studied separately, spanning feasibility and affordance issues, temporal/sequence preconditions, instruction-level logic conflicts, norm and policy violations, and multi-user disagreements. We also described a handling policy that links mismatch types to distinct interventions, including verification probes, plan adaptation, clarification with alternatives, consent/permission requests, and coordination prompts for conflicting stakeholders. 
Future work will extend the case set to richer environments and longer-horizon tasks, integrate the framework with live IoT and sensor deployments, improve uncertainty handling and threshold calibration, and strengthen norm and multi-user reasoning with more explicit role, consent, and arbitration mechanisms.

\bibliographystyle{ACM-Reference-Format}
\bibliography{sample-base}

@InProceedings{Stiber23,
  author        = "Maia Stiber and Russell H. Taylor and Chien-Ming Huang",
  title         = "On Using Social Signals to Enable Flexible Error-Aware HRI",
  booktitle     = "ACM/IEEE International Conference on Human-Robot Interaction",
  year          = 2023,
  pages         = "222--230",
  doi           = "10.1145/3568162.3576990"
}

@InProceedings{Spitale24,
  author        = "Micol Spitale and Maria Teresa Parreira and Maia Stiber and Minja Axelsson and Neval Kara and Garima Kankariya and Chien-Ming Huang and Malte F. Jung and Wendy Ju and Hatice Gunes",
  title         = "ERR@HRI 2024 Challenge: Multimodal Detection of Errors and Failures in Human-Robot Interactions",
  booktitle     = "ACM International Conference on Multimodal Interaction",
  year          = 2024,
  pages         = "652--656",
  doi = {10.1145/3678957.3689030}
}

@InProceedings{Inceoglu21,
  author        = "Arda Inceoglu and Eren Erdal Aksoy and Abdullah Cihan Ak and Sanem Sariel",
  title         = "FINO-Net: A Deep Multimodal Sensor Fusion Framework for Manipulation Failure Detection",
  booktitle     = "IEEE/RSJ International Conference on Intelligent Robots and Systems",
  year          = 2021,
  pages         = "6841--6847",
  doi           ={10.1109/IROS51168.2021.9636455}
}

@Article{Inceoglu24,
  author        = "Arda Inceoglu and Eren Erdal Aksoy and Sanem Sariel",
  title         = "Multimodal Detection and Classification of Robot Manipulation Failures",
  journal       = "IEEE Robotics and Automation Letters",
  volume        = 9,
  number        = 2,
  year          = 2024,
  pages         = "1396--1403",
  doi={10.1109/LRA.2023.3346270}
}

@InProceedings{Ichter23a,
  title = 	 {Do As I Can, Not As I Say: Grounding Language in Robotic Affordances},
  author =       {Ichter, Brian and Brohan, Anthony and Chebotar, Yevgen and Finn, Chelsea and Hausman, Karol and Herzog, Alexander and Ho, Daniel and Ibarz, Julian and Irpan, Alex and Jang, Eric and Julian, Ryan and Kalashnikov, Dmitry and Levine, Sergey and Lu, Yao and Parada, Carolina and Rao, Kanishka and Sermanet, Pierre and Toshev, Alexander T and Vanhoucke, Vincent and Xia, Fei and Xiao, Ted and Xu, Peng and Yan, Mengyuan and Brown, Noah and Ahn, Michael and Cortes, Omar and Sievers, Nicolas and Tan, Clayton and Xu, Sichun and Reyes, Diego and Rettinghouse, Jarek and Quiambao, Jornell and Pastor, Peter and Luu, Linda and Lee, Kuang-Huei and Kuang, Yuheng and Jesmonth, Sally and Joshi, Nikhil J. and Jeffrey, Kyle and Ruano, Rosario Jauregui and Hsu, Jasmine and Gopalakrishnan, Keerthana and David, Byron and Zeng, Andy and Fu, Chuyuan Kelly},
  booktitle = 	 {Proceedings of The 6th Conference on Robot Learning},
  pages         = 	 {287--318},
  year          = 	 {2023},
  volume        = 	 {205},
  series        = 	 {Proceedings of Machine Learning Research},
  month         = 	 {14--18 Dec},
  publisher     =    {PMLR},
}

@InProceedings{Zitkovich23a,
  title         = 	 {RT-2: Vision-Language-Action Models Transfer Web Knowledge to Robotic Control},
  author        =       {Zitkovich, Brianna and Yu, Tianhe and Xu, Sichun and Xu, Peng and Xiao, Ted and Xia, Fei and Wu, Jialin and Wohlhart, Paul and Welker, Stefan and Wahid, Ayzaan and Vuong, Quan and Vanhoucke, Vincent and Tran, Huong and Soricut, Radu and Singh, Anikait and Singh, Jaspiar and Sermanet, Pierre and Sanketi, Pannag R. and Salazar, Grecia and Ryoo, Michael S. and Reymann, Krista and Rao, Kanishka and Pertsch, Karl and Mordatch, Igor and Michalewski, Henryk and Lu, Yao and Levine, Sergey and Lee, Lisa and Lee, Tsang-Wei Edward and Leal, Isabel and Kuang, Yuheng and Kalashnikov, Dmitry and Julian, Ryan and Joshi, Nikhil J. and Irpan, Alex and Ichter, Brian and Hsu, Jasmine and Herzog, Alexander and Hausman, Karol and Gopalakrishnan, Keerthana and Fu, Chuyuan and Florence, Pete and Finn, Chelsea and Dubey, Kumar Avinava and Driess, Danny and Ding, Tianli and Choromanski, Krzysztof Marcin and Chen, Xi and Chebotar, Yevgen and Carbajal, Justice and Brown, Noah and Brohan, Anthony and Arenas, Montserrat Gonzalez and Han, Kehang},
  booktitle     = 	 {Proceedings of The 7th Conference on Robot Learning},
  pages         = 	 {2165--2183},
  year          = 	 {2023},
  volume        = 	 {229},
  publisher     =    {PMLR},
  url = 	 {https://proceedings.mlr.press/v229/zitkovich23a.html},
  }

@inproceedings{Kontogiorgos20,
    author      = {Kontogiorgos, Dimosthenis and Pereira, Andre and Sahindal, Boran and van Waveren, Sanne and Gustafson, Joakim},
    title       = {Behavioural Responses to Robot Conversational Failures},
    year        = {2020},
    isbn        = {9781450367462},
    doi         = {10.1145/3319502.3374782},
    booktitle   = {Proceedings of the 2020 ACM/IEEE International Conference on Human-Robot Interaction},
    pages       = {53–62},
    numpages    = {10},
    }

@INPROCEEDINGS{Esterwood22,
      author    ={Esterwood, Connor and Robert, Lionel P.},
      booktitle ={2022 17th ACM/IEEE International Conference on Human-Robot Interaction (HRI)}, 
      title     ={Having the Right Attitude: How Attitude Impacts Trust Repair in Human—Robot Interaction}, 
      year      ={2022},
      volume    ={},
      number    ={},
      pages     ={332-341},
      doi       ={10.1109/HRI53351.2022.9889535}}

@InProceedings{Kontogiorgos21,
  author    = {Kontogiorgos, Dimosthenis and Tran, Minh and Gustafson, Joakim and Soleymani, Mohammad},
  title     = {A Systematic Cross-Corpus Analysis of Human Reactions to Robot Conversational Failures},
  booktitle = {Proceedings of the 2021 International Conference on Multimodal Interaction},
  year      = {2021},
  pages     = {112--120},
  doi = {10.1145/3462244.3479887}
}

@InProceedings{Ruddy25,
  author        = "Patamia, Rutherford Agbeshi and Dinh, Ha Pham Thien and Liu, Ming and Cosgun, Akansel",
  title         = "Beyond Technical Failures: Multimodal Time-Series Modelling for Detecting Social Breakdowns and User Repair Attempts in Human-Robot Interaction",
  booktitle     = "ACM International Conference on Multimedia",
  year          = 2025,
  pages         = {14136–14142},
  doi = {10.1145/3746027.3762074}
}

@ARTICLE{Zhu21,
    
AUTHOR={Zhu, Fan  and Wang, Liangliang  and Wen, Yilin  and Yang, Lei  and Pan, Jia  and Wang, Zheng  and Wang, Wenping },
           
TITLE={Failure Handling of Robotic Pick and Place Tasks With Multimodal Cues Under Partial Object Occlusion},
          
JOURNAL={Frontiers in Neurorobotics},
          
VOLUME={Volume 15},
  
YEAR={2021},
  
DOI={10.3389/fnbot.2021.570507},
  
ISSN={1662-5218},
}

@Article{Ji22,
  author        = "Ji, Tianchen and Sivakumar, Arun Narenthiran and Chowdhary, Girish and Driggs-Campbell, Katherine",
  title         = "Proactive Anomaly Detection for Robot Navigation With Multi-Sensor Fusion",
  journal       = "IEEE Robotics and Automation Letters",
  volume        = 7,
  number        = 2,
  year          = 2022,
  pages         = "4975--4982",
  doi={10.1109/LRA.2022.3153989}
}

@InProceedings{Mitrevski21,
  author        = "Mitrevski, Alex and Plöger, Paul G. and Lakemeyer, Gerhard",
  title         = "Robot Action Diagnosis and Experience Correction by Falsifying Parameterised Execution Models",
  booktitle     = "IEEE International Conference on Robotics and Automation",
  year          = 2021,
  pages         = "11025--11031",
  doi = {10.1109/ICRA48506.2021.9561710}
}

@Article{Pasricha22,
  author={Pasricha, Anuj and Tung, Yi-Shiuan and Hayes, Bradley and Roncone, Alessandro},
  journal={IEEE Robotics and Automation Letters}, 
  title={PokeRRT: Poking as a Skill and Failure Recovery Tactic for Planar Non-Prehensile Manipulation}, 
  year={2022},
  volume={7},
  number={2},
  pages={4480-4487},
  doi={10.1109/LRA.2022.3148442}
}

@InProceedings{Brohan23,
  author        = "Brohan, Anthony and Brown, Noah and Carbajal, Justice and Chebotar, Yevgen and Dabis, Joseph and Finn, Chelsea and Gopalakrishnan, Keerthana and Hausman, Karol and Herzog, Alexander and Hsu, Jasmine and Ibarz, Julian and Ichter, Brian and Irpan, Alex and Jackson, Tomas and Jesmonth, Sally and Joshi, Nikhil and Julian, Ryan and Kalashnikov, Dmitry and Kuang, Yuheng and Leal, Isabel and Lee, Kuang-Huei and Levine, Sergey and Lu, Yao and Malla, Utsav and Manjunath, Deeksha and Mordatch, Igor and Nachum, Ofir and Parada, Carolina and Peralta, Jodilyn and Perez, Emily and Pertsch, Karl and Quiambao, Jornell and Rao, Kanishka and Ryoo, Michael S and Salazar, Grecia and Sanketi, Pannag R and Sayed, Kevin and Singh, Jaspiar and Sontakke, Sumedh and Stone, Austin and Tan, Clayton and Tran, Huong and Vanhoucke, Vincent and Vega, Steve and Vuong, Quan H and Xia, Fei and Xiao, Ted and Xu, Peng and Xu, Sichun and Yu, Tianhe and Zitkovich, Brianna",
  title         = "RT-1: Robotics Transformer for Real-World Control at Scale",
  booktitle     = "Proceedings of Robotics: Science and Systems",
  year          = 2023
}

@inproceedings{Driess23,
    author = {Driess, Danny and Xia, Fei and Sajjadi, Mehdi S. M. and Lynch, Corey and Chowdhery, Aakanksha and Ichter, Brian and Wahid, Ayzaan and Tompson, Jonathan and Vuong, Quan and Yu, Tianhe and Huang, Wenlong and Chebotar, Yevgen and Sermanet, Pierre and Duckworth, Daniel and Levine, Sergey and Vanhoucke, Vincent and Hausman, Karol and Toussaint, Marc and Greff, Klaus and Zeng, Andy and Mordatch, Igor and Florence, Pete},
    title = {PaLM-E: an embodied multimodal language model},
    year = {2023},
    booktitle = {Proceedings of the 40th International Conference on Machine Learning},
    pages = {8469–8488},
    series = {ICML'23}
}

@InProceedings{Liang23,
  author        = "Liang, Jacky and Huang, Wenlong and Xia, Fei and Xu, Peng and Hausman, Karol and Ichter, Brian and Florence, Pete and Zeng, Andy",
  title         = "Code as Policies: Language Model Programs for Embodied Control",
  booktitle     = "IEEE International Conference on Robotics and Automation",
  year          = 2023,
  pages         ={9493-9500},
  doi={10.1109/ICRA48891.2023.10160591}
}

@InProceedings{Huang22,
  author        = "Huang, Wenlong and Xia, Fei and Xiao, Ted and Chan, Harris and Liang, Jacky and Florence, Pete and Zeng, Andy and Tompson, Jonathan and Mordatch, Igor and Chebotar, Yevgen and Sermanet, Pierre and Jackson, Tomas and Brown, Noah and Luu, Linda and Levine, Sergey and Hausman, Karol and Ichter, Brian",
  title         = "Inner Monologue: Embodied Reasoning through Planning with Language Models",
  booktitle     = "Conference on Robot Learning",
  year          = 2023,
  pages         =  {1769--1782},
  volume        =  {205}
}

@article{Jackson22,
author = {Jackson, Ryan Blake and Williams, Tom},
title = {Enabling Morally Sensitive Robotic Clarification Requests},
year = {2022},
publisher = {Association for Computing Machinery},
volume = {11},
number = {2},
url = {https://doi.org/10.1145/3503795},
doi = {10.1145/3503795},
journal = {ACM Transactions on Human-Robot Interaction},
articleno = {17},
pages = {1-18},
}

@Article{ZhangIJHCS23,
    title = {“Sorry, it was my fault”: Repairing trust in human-robot interactions},
    journal = {International Journal of Human-Computer Studies},
    volume = {175},
    pages = {103031},
    year = {2023},
    issn = {1071-5819},
    doi = {https://doi.org/10.1016/j.ijhcs.2023.103031},
    author = {Xinyi Zhang and Sun Kyong Lee and Whani Kim and Sowon Hahn},
}

@Article{ZhangIJSR23,
  author        = "Zhang, Xinyi and Lee, Sun Kyong and Maeng, Hoyoung and Hahn, Sowon",
  title         = "Effects of Failure Types on Trust Repairs in Human--Robot Interactions",
  journal       = "International Journal of Social Robotics",
  volume        = 15,
  year          = 2023,
  pages         = "1619--1635",
  DOI = "10.1007/s12369-023-01059-0"
}

@Article{Esterwood23,
  author        = "Esterwood, Connor and Robert, Lionel P.",
  title         = "Three Strikes and you are out! The impacts of multiple human–robot trust violations and repairs on robot trustworthiness",
  journal       = "Computers in Human Behavior",
  volume        = 142,
  year          = 2023,
  pages         = "107658",
  DOI = "10.1016/j.chb.2023.107658"
}

@InProceedings{Nesset23,
  author    = {Nesset, Birthe and Romeo, Marta and Rajendran, Gnanathusharan and Hastie, Helen},
  title     = {Robot Broken Promise? Repair strategies for mitigating loss of trust for repeated failures},
  booktitle = {IEEE International Conference on Robot and Human Interactive Communication},
  year      = {2023},
  pages     = {1389--1395},
  doi={10.1109/RO-MAN57019.2023.10309558}
}

@ARTICLE{Honig18,
    
    AUTHOR={Honig, Shanee  and Oron-Gilad, Tal },
               
    TITLE={Understanding and Resolving Failures in Human-Robot Interaction: Literature Review and Model Development},
              
    JOURNAL={Frontiers in Psychology},
              
    VOLUME={Volume 9},
      
    YEAR={2018},
      
    DOI={10.3389/fpsyg.2018.00861},
      
    ISSN={1664-1078},
}

@InProceedings{Tolmeijer20,
  author    = {Tolmeijer, Suzanne and Weiss, Astrid and Hanheide, Marc and Lindner, Felix and Powers, Thomas M and Dixon, Clare and Tielman, Myrthe L.},
  title     = {Taxonomy of Trust-Relevant Failures and Mitigation Strategies},
  booktitle = {ACM/IEEE International Conference on Human-Robot Interaction},
  year      = {2020},
  pages     = {3--12},
  numpages = {10},
  doi = {10.1145/3319502.3374793}
}

@Article{Deits13,
  author        = "Deits, Robin and Tellex, Stefanie and Thaker, Pratiksha and Simeonov, Dimitar and Kollar, Thomas and Roy, Nicholas",
  title         = "Clarifying Commands with Information-Theoretic Human-Robot Dialog",
  journal       = "Journal of Human-Robot Interaction",
  volume        = 2,
  number        = 2,
  year          = 2013,
  pages         = "58--79",
  doi = {10.5898/JHRI.2.2.Deits}
}

@InProceedings{Tellex11,
  author        = "Stefanie Tellex and Thomas Kollar and Steven Dickerson and Matthew R. Walter and Ashis Gopal Banerjee and Seth Teller and Nicholas Roy",
  title         = "Understanding Natural Language Commands for Robotic Navigation and Mobile Manipulation",
  booktitle     = "Proceedings of the AAAI Conference on Artificial Intelligence",
  volume        = 25,
  number        = 1,
  year          = 2011,
  pages         = "1507--1514",
  DOI = "10.1609/aaai.v25i1.7979"
}

@article{Tian21,
author = {Tian, Leimin and Oviatt, Sharon},
title = {A Taxonomy of Social Errors in Human-Robot Interaction},
year = {2021},
volume = {10},
number = {2},
url = {https://doi.org/10.1145/3439720},
doi = {10.1145/3439720},
journal = {ACM Transactions on Human-Robot Interaction},
articleno = {13},
numpages = {32},
}

@Article{Mirnig17,
  author        = "Nicole Mirnig and Gerald Stollnberger and Markus Miksch and Susanne Stadler and Manuel Giuliani and Manfred Tscheligi",
  title         = "To Err Is Robot: How Humans Assess and Act toward an Erroneous Social Robot",
  journal       = "Frontiers in Robotics and AI",
  volume        = 4,
  year          = 2017,
  DOI = "10.3389/frobt.2017.00021"
}

@InProceedings{Reig21,
  author        = "Reig, Samantha and Carter, Elizabeth J. and Fong, Terrence and Forlizzi, Jodi and Steinfeld, Aaron",
  title         = "Flailing, Hailing, Prevailing: Perceptions of Multi-Robot Failure Recovery Strategies",
  booktitle     = "ACM/IEEE International Conference on Human-Robot Interaction",
  year          = 2021,
  pages         = "158--167",
  doi = {10.1145/3434073.3444659}
}

@InProceedings{BobuHri21,
  author        = "Bobu, Andreea and Peng, Andi and Agrawal, Pulkit and Shah, Julie A. and Dragan, Anca D.",
  title         = "Aligning Human and Robot Representations",
  booktitle     = "ACM/IEEE International Conference on Human-Robot Interaction",
  year          = 2024,
  pages         = "42--54",
  doi           = "10.1145/3610977.3634987"
}

@InProceedings{Denning09,
  author        = "Denning, Tamara and Matuszek, Cynthia and Koscher, Karl and Smith, Joshua R. and Kohno, Tadayoshi",
  title         = "A Spotlight on Security and Privacy Risks with Future Household Robots: Attacks and Lessons",
  booktitle     = "Proceedings of the 11th International Conference on Ubiquitous Computing",
  year          = 2009,
  pages         = "105--114",
  doi = {10.1145/1620545.1620564}
}

@InProceedings{Wachowiak24,
  author        = "Wachowiak, Lennart and Tisnikar, Peter and Coles, Andrew and Canal, Gerard and Celiktutan, Oya",
  title         = "A Time Series Classification Pipeline for Detecting Interaction Ruptures in HRI Based on User Reactions",
  booktitle     = "Proceedings of the 26th International Conference on Multimodal Interaction",
  year          = 2024,
  pages         = "657--665",
  doi = {10.1145/3678957.3688386}
}

@InProceedings{Pramanick24,
  author        = "Pradip Pramanick and Silvia Rossi",
  title         = "PRISCA at ERR@HRI 2024: Multimodal Representation Learning for Detecting Interaction Ruptures in HRI",
  booktitle     = "Proceedings of the 26th International Conference on Multimodal Interaction",
  year          = 2024,
  pages         = "666--670",
doi = {10.1145/3678957.3688387}
}

@InProceedings{Janssens24,
  author        = "Janssens, Ruben and Verhelst, Eva and De Coster, Mathieu",
  title         = "Predicting Errors and Failures in Human-Robot Interaction from Multi-Modal Temporal Data",
  booktitle     = "Proceedings of the 26th International Conference on Multimodal Interaction",
  year          = 2024,
  pages         = "671--676",
doi = {10.1145/3678957.3688388}
}

@InProceedings{LiRoss23,
  author        = "Li, Na and Ross, Robert",
  title         = "Hmm, You Seem Confused! Tracking Interlocutor Confusion for Situated Task-Oriented HRI",
  booktitle     = "ACM/IEEE International Conference on Human-Robot Interaction",
  year          = 2023,
  pages         = "142--151",
doi = {10.1145/3568162.3576999}
}

@Article{LiCourtneyRoss25,
  author        = "Na Li and Jane Courtney and Robert Ross",
  title         = "HRI-confusion: A multimodal dataset for modelling and detecting user confusion in situated human-robot interaction",
  journal       = "Data in Brief",
  volume        = 62,
  year          = 2025,
  pages         = "112047",
  DOI = "10.1016/j.dib.2025.112047"
}

@InProceedings{LeMasurier24,
  author        = "LeMasurier, Gregory and Gautam, Alvika and Han, Zhao and Crandall, Jacob W. and Yanco, Holly A.",
  title         = "Reactive or Proactive? How Robots Should Explain Failures",
  booktitle     = "ACM/IEEE International Conference on Human-Robot Interaction",
  year          = 2024,
  pages         = "413--422",
  doi = {10.1145/3610977.3634963}
}

@Article{Kaelbling98,
  author        = "Leslie Pack Kaelbling and Michael L. Littman and Anthony R. Cassandra",
  title         = "Planning and Acting in Partially Observable Stochastic Domains",
  journal       = "Artificial Intelligence",
  volume        = 101,
  year          = 1998,
  pages         = "99--134",
  DOI = "10.1016/S0004-3702(98)00023-X"
}

@inproceedings{Cameron,
    author = {Cameron, Harriet R. and Castle-Green, Simon and Chughtai, Muhammad and Dowthwaite, Liz and Kucukyilmaz, Ayse and Maior, Horia A. and Ngo, Victor and Schneiders, Eike and Stahl, Bernd C.},
    title = {A Taxonomy of Domestic Robot Failure Outcomes: Understanding the impact of failure on trustworthiness of domestic robots},
    year = {2024},
    isbn = {9798400709890},
    publisher = {Association for Computing Machinery},
    url = {https://doi.org/10.1145/3686038.3686050},
    doi = {10.1145/3686038.3686050},
    booktitle = {Proceedings of the Second International Symposium on Trustworthy Autonomous Systems},
    articleno = {7},
    numpages = {14},
    series = {TAS '24}
}

@inproceedings{nogueira2024taxonomy,
    title={Don't Go Breaking My Trust: A Taxonomy of Interaction Failures for Transparent Robots},
    author={Beatriz Nogueira and Jo{\~a}o Ferreira and Hugo Sim{\~a}o and Filipa Rocha and Isabel Neto and Jo{\~a}o Guerreiro and Tiago Guerreiro},
    booktitle={Workshop on Designing Transparent and Understandable Robots},
    year={2026},
    url={https://openreview.net/forum?id=pX5ztNGY2O}
}

@article{civit2025multiuser,
  title={Multi-User Personalisation in Human-Robot Interaction: Resolving Preference Conflicts Using Gradual Argumentation},
  author={Civit, Aniol and Andriella, Antonio and Sierra, Carles and Alenyà, Guillem},
  journal={arXiv preprint arXiv:2511.03576},
  year={2025}
}

@article{10.1145/3415247,
    author = {Sebo, Sarah and Stoll, Brett and Scassellati, Brian and Jung, Malte F.},
    title = {Robots in Groups and Teams: A Literature Review},
    year = {2020},
    issue_date = {October 2020},
    publisher = {Association for Computing Machinery},
    address = {New York, NY, USA},
    volume = {4},
    number = {CSCW2},
    url = {https://doi.org/10.1145/3415247},
    doi = {10.1145/3415247},
    journal = {Proceedings of the ACM on Human-Computer Interaction},
    month = oct,
    articleno = {176},
    numpages = {36},
}

@inproceedings{Salem8520654,
    author = {Salem, Maha and Lakatos, Gabriella and Amirabdollahian, Farshid and Dautenhahn, Kerstin},
    title = {Would You Trust a (Faulty) Robot? Effects of Error, Task Type and Personality on Human-Robot Cooperation and Trust},
    year = {2015},
    isbn = {9781450328838},
    publisher = {Association for Computing Machinery},
    address = {New York, NY, USA},
    url = {https://doi.org/10.1145/2696454.2696497},
    doi = {10.1145/2696454.2696497},
    booktitle = {Proceedings of the Tenth Annual ACM/IEEE International Conference on Human-Robot Interaction},
    pages = {141–148},
    numpages = {8},
    location = {Portland, Oregon, USA},
    series = {HRI '15}
}

@article{bin2025survey,
  title={A survey on the visual perception of humanoid robot},
  author={Bin, Teng and Yan, Hanming and Wang, Ning and Nikoli{\'c}, Milutin N and Yao, Jianming and Zhang, Tianwei},
  journal={Biomimetic Intelligence and Robotics},
  volume={5},
  number={1},
  pages={100197},
  year={2025},
  publisher={Elsevier}
}

\appendix

\section*{Appendix}

\section{LLM Templates and a Worked Example for Stage I: Assumption extraction and hypothesis formation (\textsc{LLM-Hypothesize})}
\label{sec:llm_prompts_wsm}

\begin{quote}
\small\ttfamily\raggedright\sloppy
You are the semantic inference module in a WSM-Aware HRI system.

Goal: Externalize the user’s implicit assumptions, propose candidate mismatch types,
and request minimal verifiable evidence to confirm/refute them.

Mismatch taxonomy:
- InstructionLogic
- PerceptualAffordance
- TemporalSequence
- NormMismatch
- MultiUserConflict

Input: \\
(1) Instruction q: Put the toy into the cup. \\
(2) Interaction context u\_\{1:t\}: Single-user, no follow-up constraints or corrections. \\
(3) Robot world-state summary W\_r\^t: \\
\ \ \ - Perception facts: objects=\{toy, cup\}; toy\_size=large; cup\_opening=small; relation(toy\_size $>$ cup\_opening)=true. \\
\ \ \ - IoT/digital facts: none. \\
\ \ \ - Execution status / preconditions: toy not grasped yet; cup is reachable. \\
\ \ \ - Roles/policies/norms: none. \\
\ \ \ - Multi-user info (if any): none.

Output in JSON only: \\
\{\newline
\ \ "assumptions": \{\newline
\ \ \ \ "physical": ["The toy fits through the cup opening.", "Placing the toy into the cup is feasible."],\newline
\ \ \ \ "temporal": ["The robot can attempt insertion after grasping the toy."],\newline
\ \ \ \ "normative": [],\newline
\ \ \ \ "multi\_user": []\newline
\ \ \},\newline
\ \ "candidate\_types": ["PerceptualAffordance"],\newline
\ \ "evidence\_requests": [\newline
\ \ \ \ \{ "check": "MeasureRelativeSize(toy, cup\_opening)", "why": "Tests whether the toy fits into the cup.", "expected\_observation": "toy\_size <= cup\_opening\_size" \}\newline
\ \ ],\newline
\ \ "tentative\_primary\_type": "PerceptualAffordance"\newline
\}\newline

Constraints: \\
- Do not finalize a mismatch type if key evidence is missing; request checks instead. \\
- Evidence requests must be verifiable (sensor probe, state-machine query, policy/role lookup,
  user/stakeholder confirmation). \\
- Keep evidence\_requests minimal (<=5).
\end{quote}

\section{LLM Prompts Templates and a Worked Example for Stage II: Evidence-grounded revision (\textsc{LLM-Revise})}
\begin{quote}
\small\ttfamily\raggedright\sloppy
You are revising a WSM hypothesis after new evidence was collected.

Input: \\
- Prior assumptions and candidate types: (JSON from Stage I) \\
- New evidence o\^t: MeasureRelativeSize(toy, cup\_opening) $\rightarrow$ toy\_diameter=7.2cm; cup\_opening\_diameter=4.5cm. \\
- Updated robot state W\_r\^\{t+1\}: objects=\{toy, cup\}; relation(toy\_diameter $>$ cup\_opening\_diameter)=true; insertion\_feasible=false.

Output in JSON only: \\
\{\newline
\ \ "wsm": true,\newline
\ \ "primary\_type": "PerceptualAffordance",\newline
\ \ "supported\_by": ["toy\_diameter=7.2cm", "cup\_opening\_diameter=4.5cm", "toy\_diameter $>$ cup\_opening\_diameter"],\newline
\ \ "failed\_assumptions": ["The toy fits through the cup opening.", "Placing the toy into the cup is feasible."],\newline
\ \ "brief\_explanation": "The toy is larger than the cup opening (7.2cm vs 4.5cm), so inserting it into the cup is not feasible.",\newline
\ \ "recommended\_handling": "clarify"\newline
\}\newline

Rules: \\
- If multiple types remain plausible, choose the one that most directly blocks safe/appropriate execution. \\
- If NormMismatch applies, it overrides other types unless the task is physically unsafe regardless of norms. \\
- Do not add facts not present in the evidence/state.
\end{quote}

\end{document}